\documentclass[sigconf]{acmart}

\usepackage{epsfig}
\usepackage{graphicx}
\usepackage{amsmath}

\usepackage{amssymb}
\usepackage{balance}

\usepackage{bm}
\usepackage{subfig}
\usepackage[ruled,linesnumbered]{algorithm2e}
\usepackage{booktabs,multirow,array}
\usepackage{float}

\newcommand{\rmbm}[1]{ \bm{{\rm{#1}}} }
\newcommand{\aboveskip}{\setlength{\abovedisplayskip}{3pt} }
\newcommand{\belowskip}{\setlength{\belowdisplayskip}{3pt} }
\AtBeginDocument{%
  \providecommand\BibTeX{{%
    \normalfont B\kern-0.5em{\scshape i\kern-0.25em b}\kern-0.8em\TeX}}}

\copyrightyear{2020} 
\acmYear{2020} 
\setcopyright{acmlicensed}\acmConference[MM '20]{Proceedings of the 28th ACM International Conference on Multimedia}{October 12--16, 2020}{Seattle, WA, USA}
\acmBooktitle{Proceedings of the 28th ACM International Conference on Multimedia (MM '20), October 12--16, 2020, Seattle, WA, USA}
\acmPrice{15.00}
\acmDOI{10.1145/3394171.3413763}
\acmISBN{978-1-4503-7988-5/20/10}

\begin{document}
\fancyhead{}
\pagestyle{plain}

\title{Nighttime Dehazing with a Synthetic Benchmark}


\author{Jing Zhang}
\affiliation{%
  \institution{UBTECH Sydney Artificial Intelligence Centre, \\The University of Sydney, Sydney, Australia}
}

\author{Yang Cao}
\affiliation{%
  \institution{University of Science and Technology of China}
  \city{Hefei}
  \country{China}
}

\author{Zheng-Jun Zha}
\affiliation{%
  \institution{University of Science and Technology of China}
  \city{Hefei}
  \country{China}
}

\author{Dacheng Tao}
\affiliation{%
  \institution{UBTECH Sydney Artificial Intelligence Centre, \\The University of Sydney, Sydney, Australia}
}



\begin{abstract}
  Increasing the visibility of nighttime hazy images is challenging because of uneven illumination from active artificial light sources and haze absorbing/scattering. The absence of large-scale benchmark datasets hampers progress in this area. To address this issue, we propose a novel synthetic method called 3R to simulate nighttime hazy images from daytime clear images, which first reconstructs the scene geometry, then simulates the light rays and object reflectance, and finally renders the haze effects. Based on it, we generate realistic nighttime hazy images by sampling real-world light colors from a prior empirical distribution. Experiments on the synthetic benchmark show that the degrading factors jointly reduce the image quality. To address this issue, we propose an optimal-scale maximum reflectance prior to disentangle the color correction from haze removal and address them sequentially. Besides, we also devise a simple but effective learning-based baseline which has an encoder-decoder structure based on the MobileNet-v2 backbone. Experiment results demonstrate their superiority over state-of-the-art methods in terms of both image quality and runtime. Both the dataset and source code will be available at \url{https://github.com/chaimi2013/3R}.
\end{abstract}



\begin{teaserfigure}
  \includegraphics[width=1\linewidth]{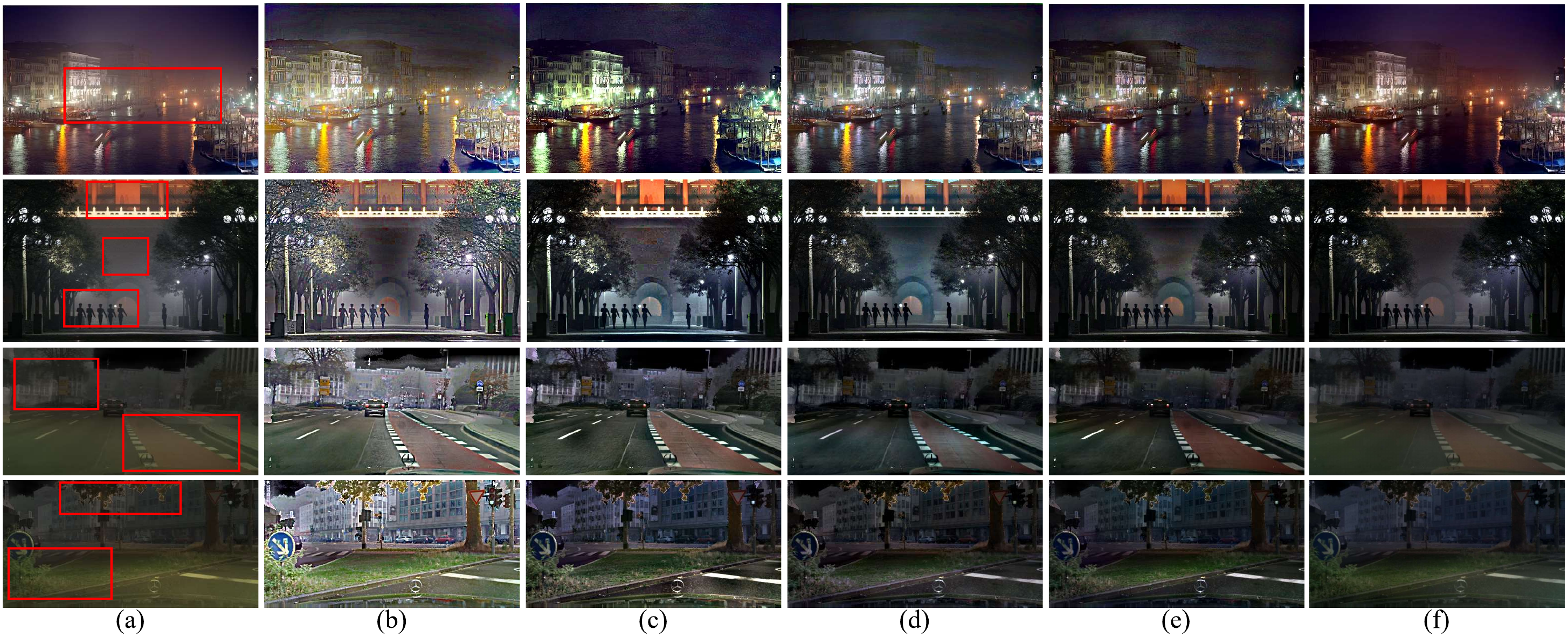}
  \captionof{figure}{(a) Nighttime hazy images. (b) NDIM \cite{zhang2014nighttime}. (c) GS \cite{li2015nighttime}. (d) MRP \cite{zhang2017fast}. (e) Our OSFD. (f) Our ND-Net.}
  \label{fig:banner}
\end{teaserfigure}

\maketitle

\section{Introduction}
In contrast to daytime imaging conditions, where the illumination is dominated by global uniform atmospheric light, nighttime illumination is mainly from active, synthetic light sources such as street and neon lights. These lights are located at different positions, have limited illumination range, and produce diverse colors, resulting in a low-visibility image with uneven illumination and color cast. If the conditions are hazy, image visibility will be even worse, since haze degrades image contrast through absorption and scattering (see examples in Figure~\ref{fig:banner}(a)). Removing color cast and haze is crucial for increasing the visibility of nighttime hazy images, which, however, is very challenging due to its ill-posed nature.

To address this problem, Pei and Lee first proposed a color-transfer technique to convert a nighttime hazy image into a grayish one by referring to the color statistics of a reference daytime image \cite{pei2012nighttime}. While reducing the color cast, it also changed the original color distribution and generated unrealistic results with color artifacts \cite{zhang2017fast}. Zhang $et~al.$ proposed a model-based method to correct the color cast \cite{zhang2014nighttime} and remove haze using the dark channel prior \cite{he2009single}. However, the light compensation added extra color cast that affected the subsequent color correction (Figure~\ref{fig:banner}(b)). Li $et~al.$ proposed a layer separation algorithm to remove the glow of light sources in the input image \cite{li2015nighttime}. While producing better haze-free images than previous methods, it also generated some color artifacts around the light sources and amplified noise (Figure~\ref{fig:banner}(c)). Recently, Zhang $et~al.$ proposed a statistical prior, the maximum reflectance prior (MRP), to help remove the color cast and then dehaze \cite{zhang2017fast}. However, MRP was used within each pixel's fixed-sized neighborhood and could not adapt to diverse local statistics such as large areas of the monochromatic lawn (Figure~\ref{fig:banner}(d)).

Less progress has been made in nighttime dehazing than daytime dehazing, especially in the context of deep learning. For example, many convolutional neural networks (CNN)-based methods have been proposed for daytime dehazing \cite{TIP_2016_Cai, ren2016single, ren2016single, li2017aod, ren2018gated, zhang2018densely, yang2018proximal, zhang2018fully, li2019lap, Deng_2019_ICCV, liu2019griddehazenet, zhang2019famed,Wang_2020_CVPR}. The benefit of the strong representation capacity of CNN relies on large-scale training data. Even when trained on synthetic images, they show good generalization ability to real-world daytime hazy images. However, this is not the case for nighttime hazy scenarios. First, models trained on synthetic daytime hazy images do not generalize well to nighttime hazy images due to the illumination discrepancy. Second, synthesizing nighttime hazy images is not straightforward due to the spatially variant illumination and its entanglement with haze scattering and absorption.

To address this issue, we propose a novel synthetic method to simulate nighttime hazy images from daytime clear images called 3R. We first conduct an empirical study on real-world light colors to obtain a prior distribution. Then, we use 3R to synthesize realistic nighttime hazy images by sampling real-world light colors. Specifically, it first reconstructs the scene geometry, then simulates the light rays and object reflectance, and finally renders the haze effects. We build a new synthetic dataset to benchmark state-of-the-art (SOTA) methods. As noted above, existing methods suffer from the entanglement of both degrading factors, i.e., active light sources and haze. To address this issue, we propose an optimal-scale maximum reflectance prior (OS-MRP) to disentangle the color correction from haze removal and address them sequentially (Figure~\ref{fig:banner}(e)). Also, we devise a new CNN baseline model with an encoder-decoder structure, which also achieves good performance at high computational efficiency (Figure~\ref{fig:banner}(f)). The main contributions of this paper are:

$\bullet$ We propose a novel synthetic method to establish a benchmark and evaluate SOTA nighttime dehazing methods comprehensively;

$\bullet$ We derive a novel OS-MRP prior that can adapt to diverse local statistics in natural images, resulting in a computationally efficient dehazing algorithm for removing color cast and haze effectively;

$\bullet$ We devise a deep CNN model, which serves as a strong baseline and achieves good performance for nighttime dehazing;

$\bullet$ Extensive experiments on synthetic datasets and real-world images demonstrate the superiority of the proposed methods in terms of both image quality and runtime.

\section{Related work}
\label{sec:relatedwork}
\begin{equation}
\begin{aligned}
\rmbm{I}\left( \rmbm{x} \right) &= {L}\left( \rmbm{x} \right)\rmbm{\eta} \left( \rmbm{x} \right)\rmbm{R}\left( \rmbm{x} \right)t\left( \rmbm{x} \right) + {L}\left( \rmbm{x} \right)\rmbm{\eta} \left( \rmbm{x} \right)\left( {1 - t\left( \rmbm{x} \right)} \right)\\
& \buildrel \Delta \over = \rmbm{J}\left( \rmbm{x} \right)\rmbm{\eta} \left( \rmbm{x} \right)t\left( \rmbm{x} \right) + {L}\left( \rmbm{x} \right)\rmbm{\eta} \left( \rmbm{x} \right)\left( {1 - t\left( \rmbm{x} \right)} \right),
\label{eq:imagingModel}
\end{aligned}
\end{equation}
The nighttime imaging model can be formulated as above by referring to \cite{zhang2017fast}. Here, $\rmbm{I}\left( \rmbm{x} \right)$ is the nighttime hazy image, ${L}\left( \rmbm{x} \right)$ and $\rmbm{\eta}\left( \rmbm{x} \right)$ are the illuminance and color cast from active light sources, $\rmbm{R}\left( \rmbm{x} \right)$ is the reflectance, $t\left( \rmbm{x} \right)$ is the haze transmission, and $\rmbm{J}\left( \rmbm{x} \right)$ is the nighttime clear image:
\begin{equation}
\rmbm{J}\left( \rmbm{x} \right) \buildrel \Delta \over = \rmbm{R}\left( \rmbm{x} \right){L}\left( \rmbm{x} \right),
\label{eq:J}
\end{equation}
where $\rmbm{x}$ is the pixel index. Recovering $\rmbm{J}$ from $\rmbm{I}$ is a typical ill-posed problem relying on the estimation of latent ${L}$, $\rmbm{\eta}$, and $t$. Previous work either used statistical priors to regularize the optimization or learned the latent variables or $\rmbm{J}$ from the training data directly.

\textbf{Image dehazing with statistical priors}: For the daytime case, $L\rmbm{\eta}$ is global constant atmospheric light. Therefore, the primary target is to estimate haze transmission. To this end, different statistical priors have been proposed \cite{he2009single, fattal2014dehazing,TIP_2015_Zhu, berman2016non} including the dark channel prior (DCP) \cite{he2009single}. However, these cannot be directly used for nighttime dehazing. For instance, DCP is no longer valid in the nighttime case where the color cast from active light sources biases the dark channel. To address this issue, previous methods have decomposed nighttime dehazing into two steps: color cast correction and haze removal \cite{pei2012nighttime, zhang2014nighttime, li2015nighttime, Ancuti2016night, zhang2017fast}. Correcting the color cast, also known as color constancy, is very well studied \cite{buchsbaum1980spatial,van2007edge,joze2012role,barron2015convolutional,barron2017fast,hu2017fc4}. However, color cast is usually a global constant which differs from the spatially variant one in the nighttime dehazing case. Li $et~al.$ estimated the local color cast using the brightest pixel prior \cite{joze2012role}  in each local patch. However, the bright pixel within a local patch may not be white, resulting in a biased estimate towards the image's intrinsic color. MRP in \cite{zhang2017fast} instead used the maximum reflectance in each channel separately, $i.e.$, they were not necessarily from the same pixel. Nevertheless, it may fail in monochromatic areas such as the lawn. By contrast, we propose a novel optimal-scale maximum reflectance prior, which can adapt to diverse local statistics.

\textbf{Image dehazing by supervised learning}: Supervised learning-based image dehazing methods \cite{TIP_2016_Cai, ren2016single, li2017aod, ren2018gated, zhang2018densely, yang2018proximal, zhang2018fully, li2019lap, Deng_2019_ICCV, zhang2019famed} need large-scale paired training samples. Daytime dehazing methods synthesize hazy samples from clear ones according to the imaging model, where the transmission is either assumed to be a constant within each local patch \cite{tang2014investigating, TIP_2016_Cai} or calculated from the scene depth \cite{ren2016single, li2017aod, li2018benchmarking, sakaridis2018semantic}. Recently, Ancuti $et~al.$ collected real hazy and corresponding haze-free images in both outdoor and indoor scenes and constructed the O-Haze (45 pairs) and IHaze (35 pairs) datasets \cite{ancuti2018ohaze, ancuti2018ihaze}. However, collecting nighttime hazy images is non-trivial due to the need for long exposures, static scenes, and illumination calibration. Instead, Zhang $et~al.$ \cite{zhang2017fast} synthesized nighttime hazy images based on the Middlebury datasets \cite{scharstein2003high} by assuming a single central light source with a constant yellow color, which limited the sampling space. By contrast, we carry out an empirical study on real-world light colors to obtain their distribution. Then, we propose a novel synthetic method by further taking the scene geometry into account. It can generate realistic nighttime hazy images by rendering the spatially variant color cast and haze simultaneously.

\section{Synthetic nighttime hazy images}
\label{sec:syn}

\subsection{Empirical study on real-world light colors}
\label{subsec:lightcolor}
\begin{figure}
\centering
\includegraphics[width=1\linewidth]{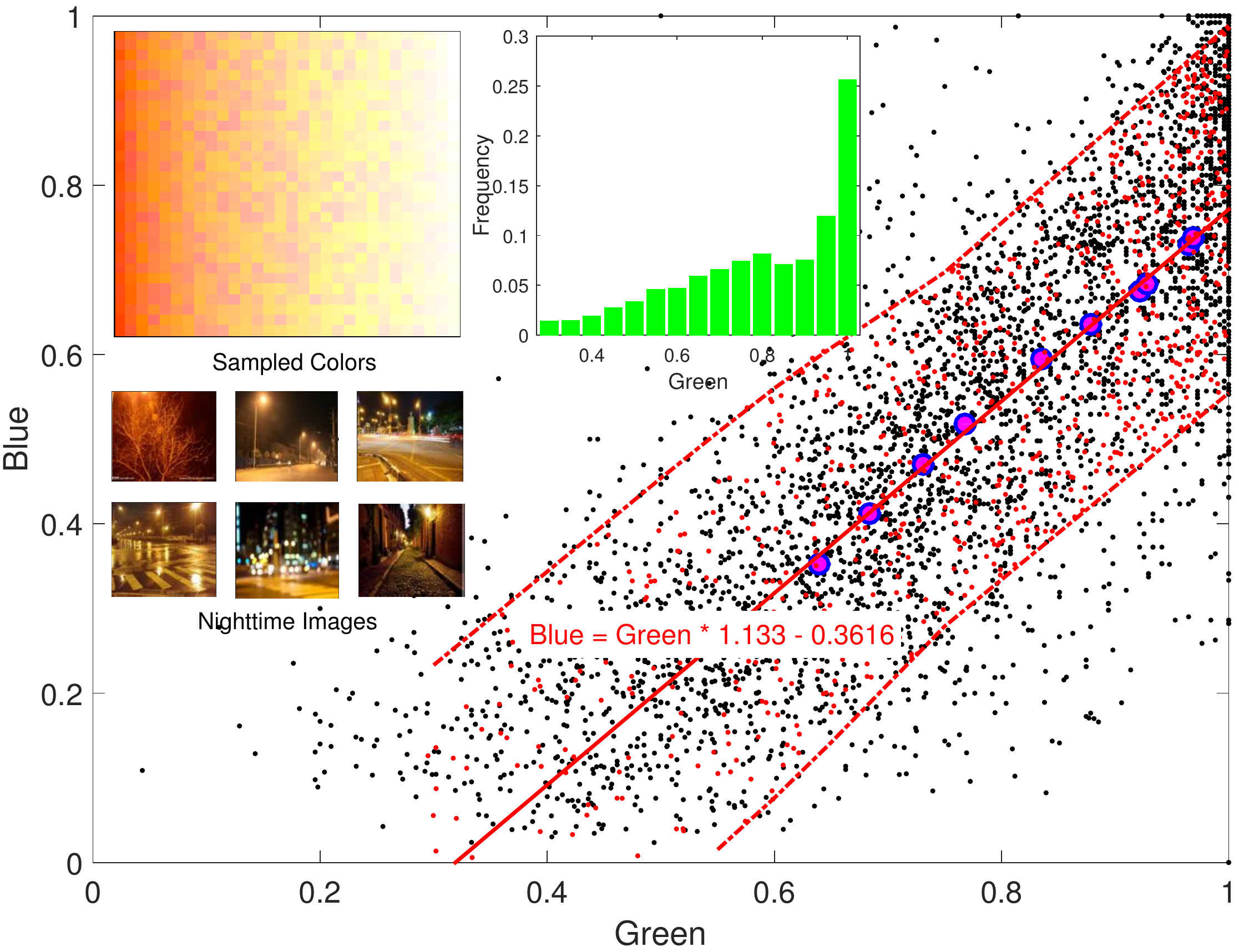}
\caption{Empirical study on the real-world light colors.}
\label{fig:lightcolor}
\end{figure}

Before synthesizing nighttime images, we must first understand real-world light colors. To this end, we collected over 1,300 real-world nighttime images from the internet using search terms such as ``nighttime road'', ''street light'', $etc$. Some examples are shown in Figure~\ref{fig:lightcolor}. Generally, the dominant illumination is from artificial active lights. Therefore, we used them to calculate the prior distribution of light colors. First, they were resized into 100x100. Then, different scales of patches were densely sampled from them, ranging from 11x11 to 100x100 pixels. At each patch, we used MRP to estimate its color cast. The results are plotted as black scattered dots in Figure~\ref{fig:lightcolor}. Note that we assume the red channel is 1 since the light colors are biased to warm colors. At each scale, we calculated the mean (center) and the standard deviation of all samples as shown in the purple circles. We then used linear regression to fit these centers and obtain a prior equation of the light colors, $i.e.$,
\begin{equation}
\aboveskip
Blue = Green \times 1.133 - 0.3616.
\belowskip
\label{eq:colorEquation}
\end{equation}
Then, we shifted the line upwards and downwards according to the standard deviations to determine the boundaries, as shown by the red dotted lines; the region bounded by them covered 98.68\% of samples. Next, we calculated the distribution of green values as shown in the green histograms. According to this distribution and Eq.~\eqref{eq:colorEquation}, we randomly sampled light colors inside the boundaries, which are plotted as red dots and visualized in the upper left corner. They are visually realistic and mimic real-world light colors.

\subsection{A novel synthetic method: 3R}
\label{subsec:3R}
Different from \cite{zhang2017fast}, where the illuminance is calculated from the light path distance via an exponential decay model, we follow the inverse-square law \cite[p.~26]{millerson2013lighting} and Lambert's cosine law \cite{basri2003lambertian} to calculate the illuminance from the light path distance, incident light direction, and surface normal direction. Therefore, we should first reconstruct scene geometry by calculating the surface normals.

\textbf{Scene reconstruction:}
Taking the Cityscapes dataset \cite{cordts2016cityscapes} as an example, given a clear image $\rmbm{R}$, semantic labels $C$, and depth map ${D}$, our target is to synthesize a realistic nighttime hazy image. First, we use SLIC \cite{achanta2012slic} to segment the semantic label into super-pixels. After calculating the world coordinates $\rmbm{x_i}$ of pixels on each super-pixel $\rmbm{z}$ based on the depth and camera parameters, a 3d plane is fitted to obtain the normal vector $\rmbm{v\left( \rmbm{z} \right )}$ and bias $m$ according to:
\begin{equation}
\aboveskip
    \rmbm{v\left( \rmbm{z} \right )} = \mathop {\min }\limits_{\rmbm{v}} \sum\limits_i {{{\left\| {{\rmbm{v}^T}{\rmbm{x_i}} + m} \right\|}^2}}.
\label{eq:planeFitting}
\belowskip
\end{equation}

\begin{figure}
\centering
\includegraphics[width=0.95\linewidth]{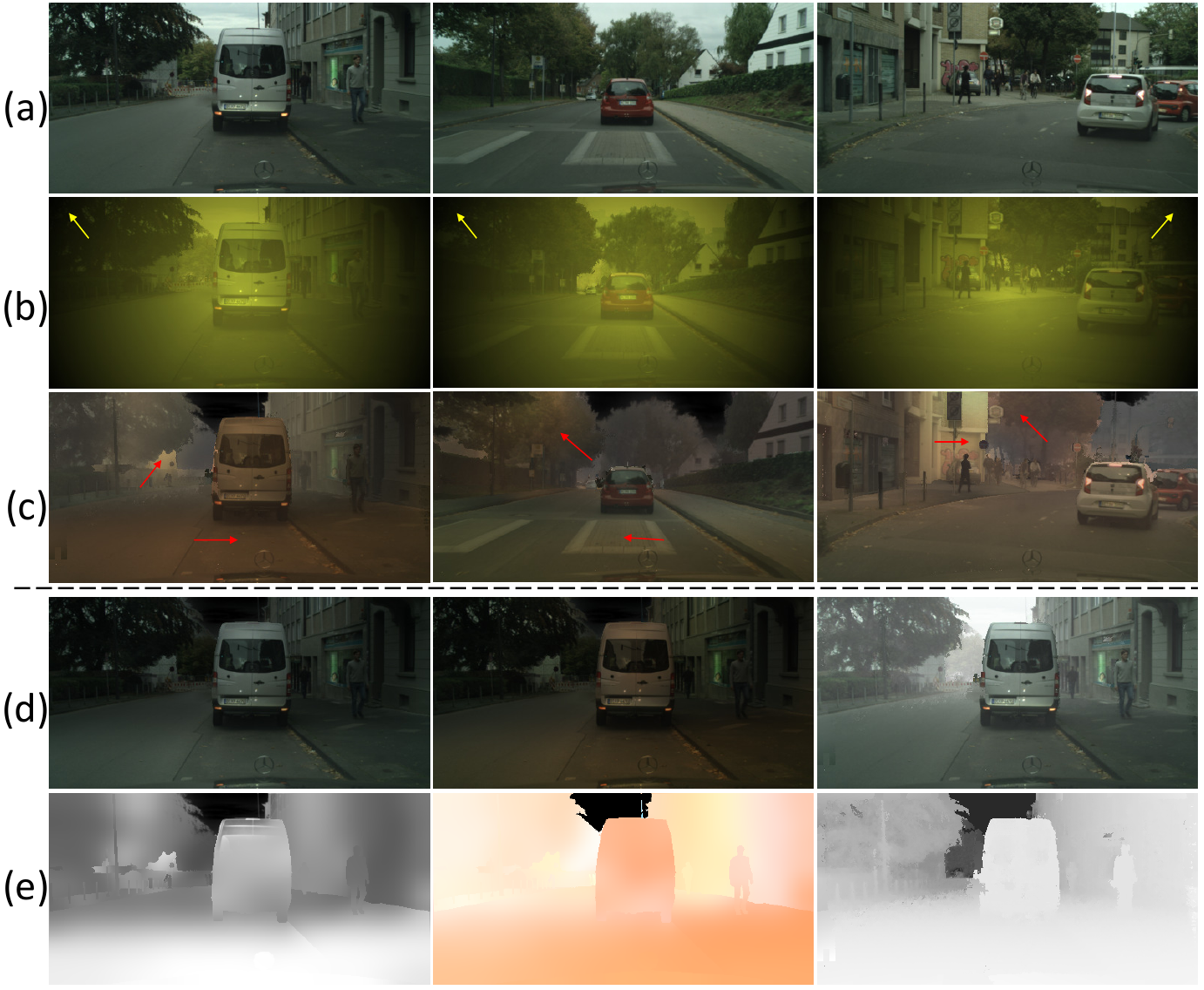}
\caption{Examples of synthetic images. (a) Daytime clear images from the Cityscapes dataset \cite{cordts2016cityscapes}. (b) Results using the method in \cite{zhang2017fast}. (c) Results using the proposed 3R. (d)-(e) Some intermediate results using 3R, $i.e.$, (d) a low-light image without/with color cast, daytime hazy image, (e) the illuminance intensity, color cast, and haze transmission.} 
\label{fig:syntheticimamgesdemo}
\end{figure}

\vspace{-5pt}
\textbf{Ray simulation:}
Then, the illuminance reaching $\rmbm{z}$ from the $k^{th}$ light can be calculated according to Lambert's cosine law \cite{basri2003lambertian}:
\begin{equation}
\rmbm{{L_k}\left( \rmbm{z} \right)} = {\rmbm{\eta _k}}\frac{{{\beta _l}}}{{d{{\left( \rmbm{z} \right)}^2}}} {\max \left( {{\rmbm{u_k}}{{\left( \rmbm{z} \right)}^T}\rmbm{v}\left( \rmbm{z} \right)} ,0 \right) }.
\label{eq:incidentlight}
\end{equation}
We use the bold $\rmbm{{L_k}}$ to denote the light intensity $L_k$ and color $\rmbm{{\eta_k}}$ together. $\beta_l$ is a parameter, $d\left( \rmbm{z} \right)$ is the distance between the $k^{th}$ light source and $\rmbm{z}$, and ${{\rmbm{u_k}}\left( \rmbm{z} \right)}$ is the incident light direction from the $k^{th}$ light to $\rmbm{z}$. Then, the illuminance from all light sources are:
\begin{equation}
\rmbm{{L}\left( \rmbm{z} \right)} = \sum\limits_k {\rmbm{{L_k}\left( \rmbm{z} \right)}}.
\label{eq:incidentlightaggr}
\end{equation}
The illuminance map is further refined using the fast guided filter \cite{he2015fast}. In our experiments, we place virtual light sources along the roadsides according to the semantic labels $C$, which are 5 meters high and spaced every 30 meters. The light colors are randomly sampled as in Section~\ref{subsec:lightcolor}.

\textbf{Rendering:}
The haze absorbing and scattering effects rely on haze transmission, which can be calculated from the scene depth according to exponential decay model \cite{sakaridis2018semantic}:
\begin{equation}
\aboveskip
t\left( \rmbm{x} \right) = {e^{ - {\beta _t}d\left( \rmbm{x} \right) }},
\label{eq:synt}
\belowskip
\end{equation}
where $\beta_t$ is the attenuation coefficient controlling the thickness of the haze. $d\left( \rmbm{x} \right)$ is the scene depth. Finally, we render the haze effect to generate the nighttime haze image by integrating the illuminance and transmission into Eq.~\eqref{eq:imagingModel}.

In this paper, we call the above method ``3R'' and summarize it in Algorithm~\ref{alg:3R}. As can be seen from the walls, roads, and cars in Figure~\ref{fig:syntheticimamgesdemo}(c), the illuminance is more realistic than Figure~\ref{fig:syntheticimamgesdemo}(b), since 3R leverages the scene geometry and real-world light colors. The haze further reduces image contrast, especially in distant regions. It is noteworthy that we mask out the sky region due to its inaccurate depth values. 3R also generates some intermediate results such as $L$, $\rmbm{\eta}$, $t$, which are worthy of further study, for example for learning disentangled representations.

\begin{algorithm}[t]
    \caption{Synthesizing Images via 3R}
    \label{alg:3R}
    \KwIn{$\rmbm{R}$, $C$, $D$, $\beta_l$, $\beta_t$, $\left\{ {{\rmbm{\eta _k}}} \right\}_{k = 1}^K$}
    \KwOut{$\rmbm{I}$}
        \SetKwFor{For}{Scene Reconstruction:}{}{}
        \For{}{
            Segment $C$ into super-pixels $\left\{ {\rmbm{z_1},...} \right\}$;\\
            Calculate the world coordinates $\rmbm{x_i}$ of pixels on $\rmbm{z}$;\\
            Calculate the normal vector $\rmbm{v}$ according to Eq.~\eqref{eq:planeFitting};\\
        }
        \SetKwFor{For}{Ray Simulation:}{}{}
        \For{}{
            Calculate the illuminance $\rmbm{L_k}$ according to Eq.~\eqref{eq:incidentlight};\\
            Aggregate the illuminance $\rmbm{L_k}$ according to Eq.~\eqref{eq:incidentlightaggr};\\
            Calculate the transmission $t$ according to Eq.~\eqref{eq:synt};\\
        }
        \SetKwFor{For}{Rendering:}{}{}
        \For{}{
            Synthesize $\rmbm{I}$ according to Eq.~\eqref{eq:imagingModel}.
        }
\end{algorithm}

\subsection{A novel synthetic benchmark}
\label{subsec:benchmark}
Following \cite{sakaridis2018semantic}, 550 clear images were selected from Cityscapes \cite{cordts2016cityscapes} to synthesize nighttime hazy images using 3R. We synthesized 5 images for each of them by changing the light positions and colors, resulting in a total of 2,750 images, called ``Nighttime Hazy Cityscapes'' (NHC). We also altered the haze density by setting $\beta_t$ to 0.005, 0.01, and 0.02, resulting in different datasets denoted NHC-L, NHC-M, and NHC-D, where ``L'', ``M'', and ``D'' represent light haze, medium haze, and dense haze. Further, we also modified the method in \cite{zhang2017fast} by changing the constant yellow light color with our randomly sampled real-world light colors described in Section~\ref{subsec:lightcolor} and synthesized images on the Middlebury (70 images) \cite{scharstein2003high} and RESIDE (8,970 images) datasets \cite{li2018benchmarking}. Similar to NHC, we augmented the Middlebury dataset by 5 times, resulting in a total of 350 images. They are denoted NHM and NHR, respectively. The statistics of these datasets are summarized in Table~\ref{tab:datasetStats} in Section~\ref{sec:experiment}.

\begin{figure}[ht]
\centering
\subfloat[]{\label{fig:mmrp_real_img}
\includegraphics[height=0.365\linewidth]{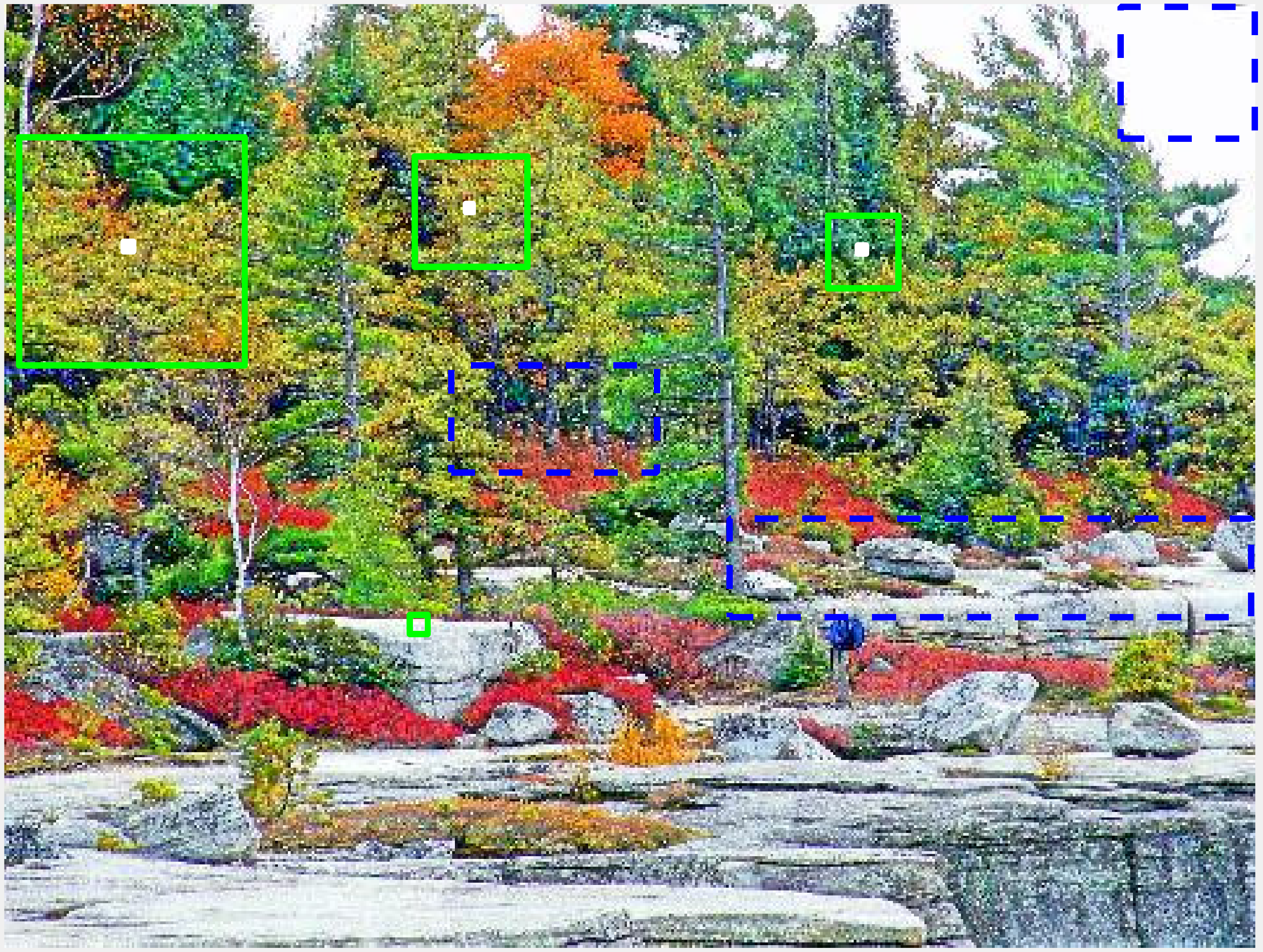}}
\hspace{0.001\linewidth}
\subfloat[]{\label{fig:mmrp_real_scaleIdx}
\includegraphics[height=0.365\linewidth]{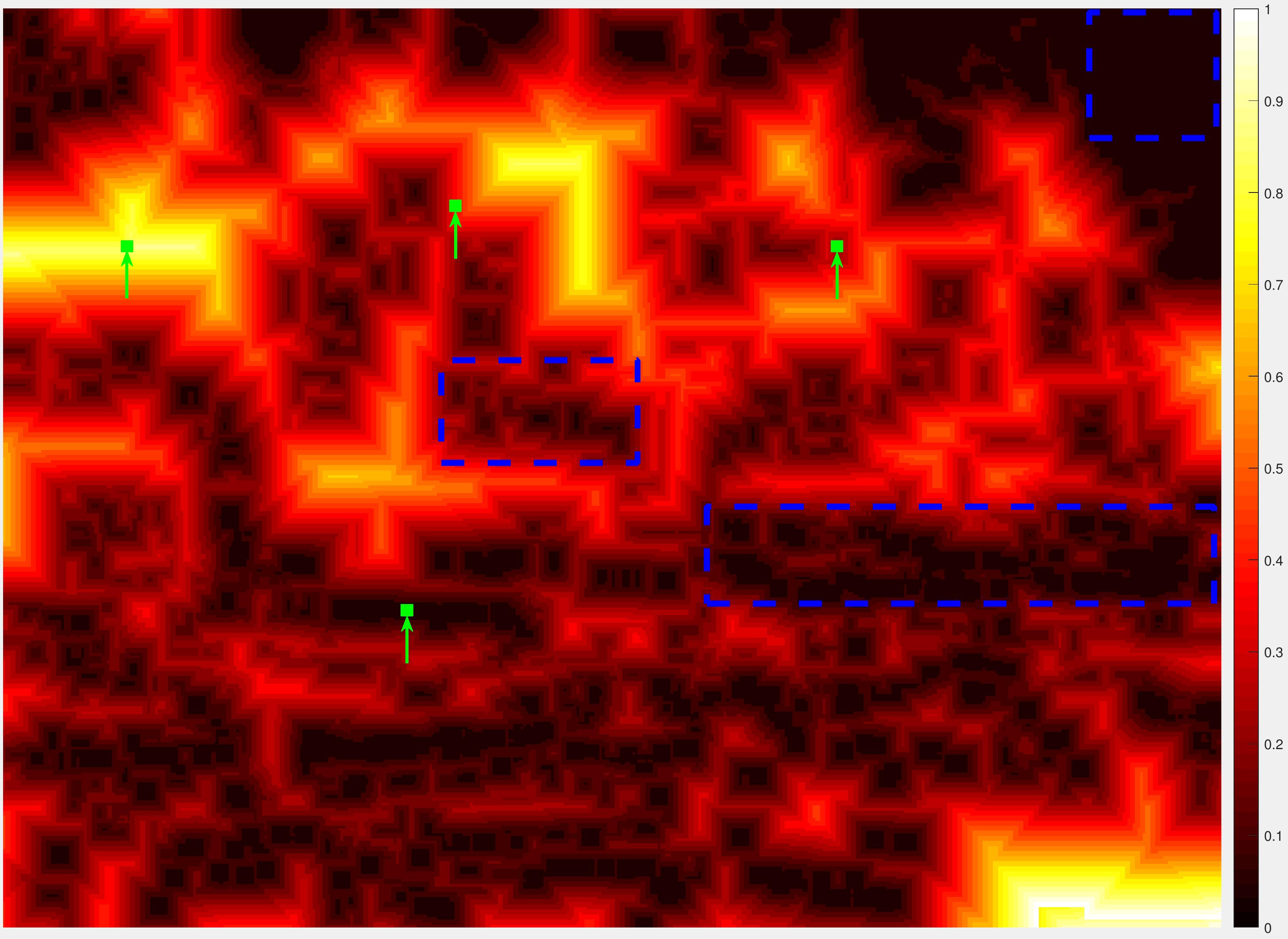}}

\caption{(a) A clear image. (b) The optimal scale map. Hot colors represent large scales. Please refer to Section~\ref{subsec:OSMRP}.}
\label{fig:osmrp_demo_real}
\end{figure}

\begin{figure*}
\centering
\subfloat[]{\label{fig:dcp}
\includegraphics[width=0.245\linewidth]{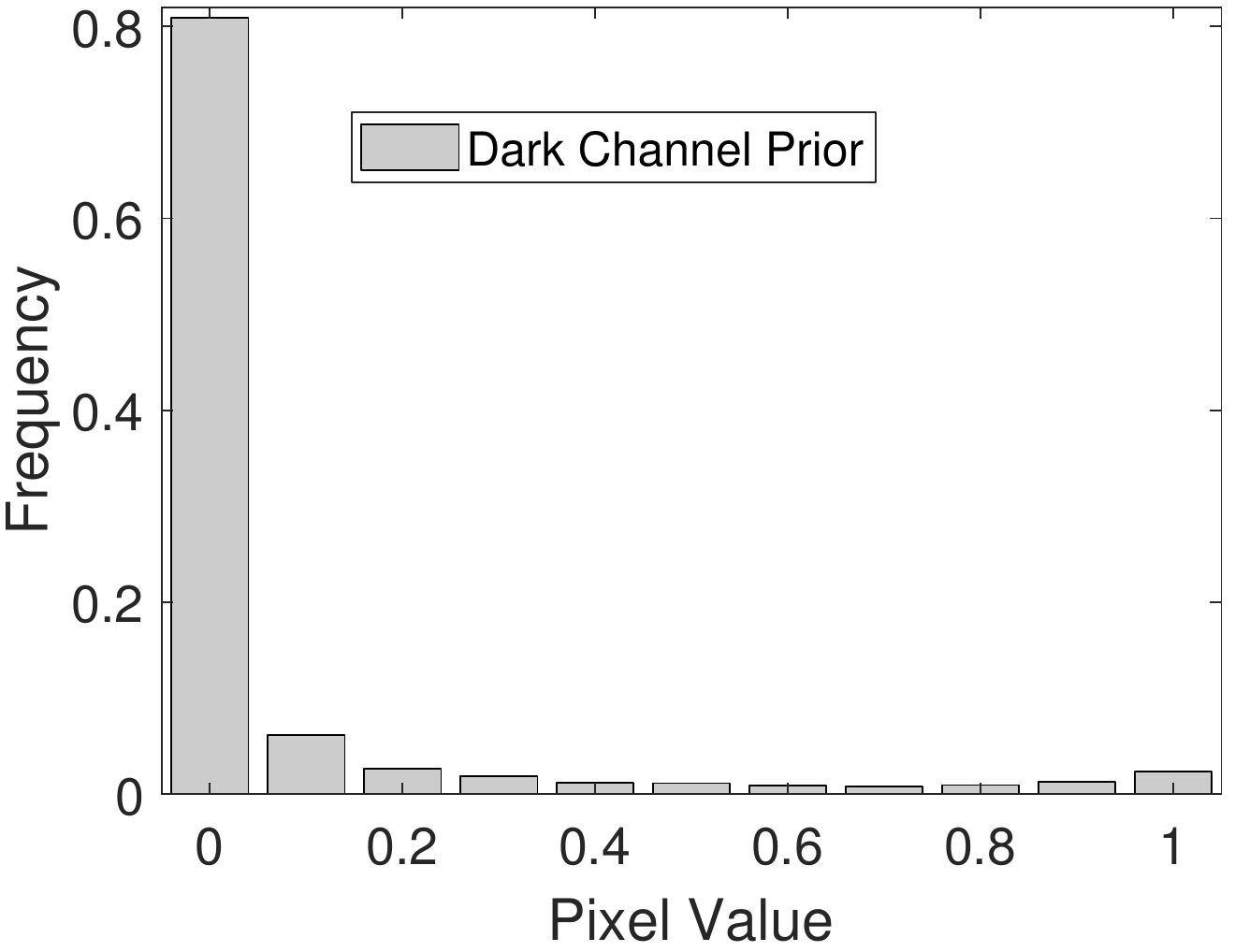}}
\hspace{0.001\linewidth}
\subfloat[]{\label{fig:mrp}
\includegraphics[width=0.245\linewidth]{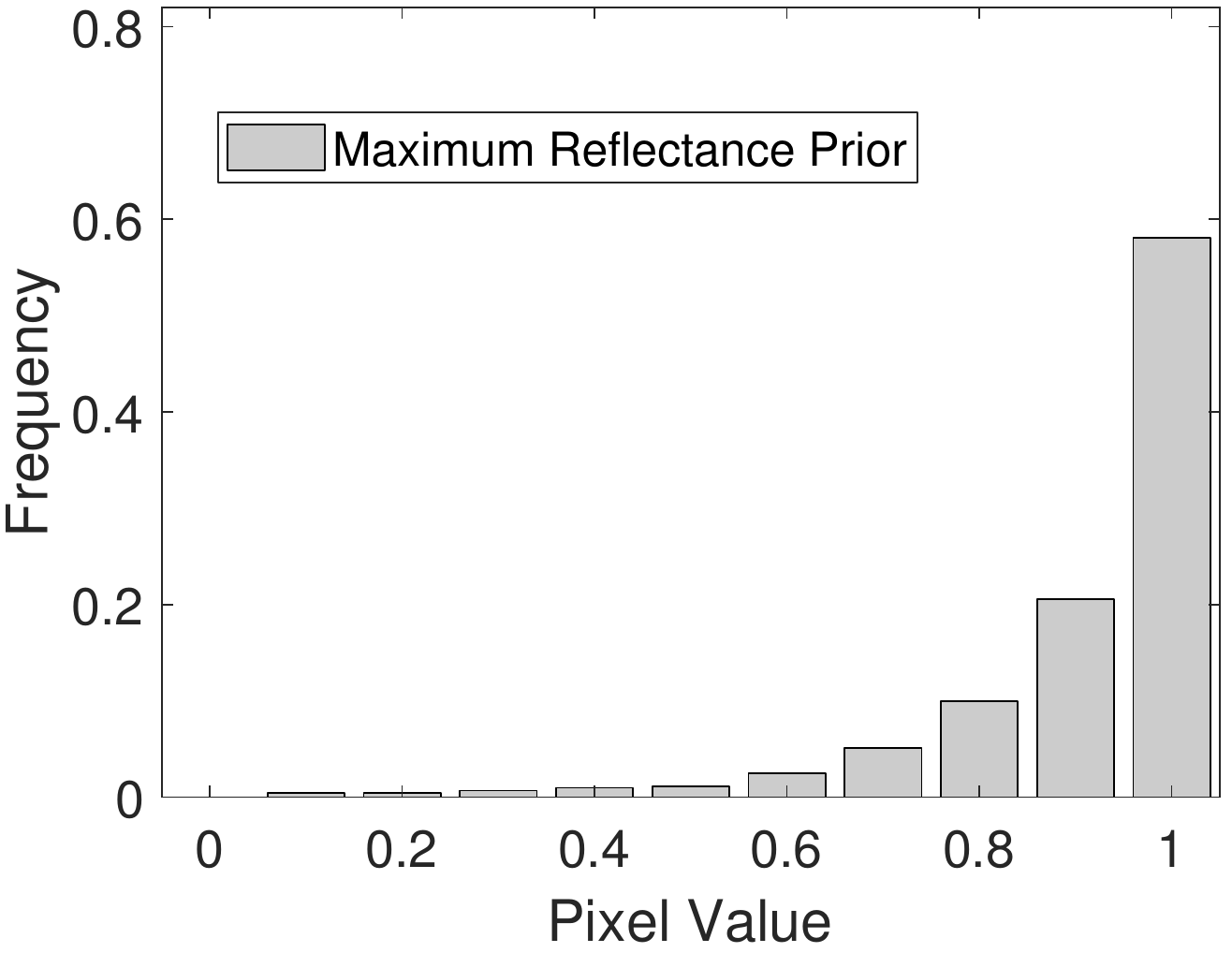}}
\hspace{0.001\linewidth}
\subfloat[]{\label{fig:mmrp}
\includegraphics[width=0.245\linewidth]{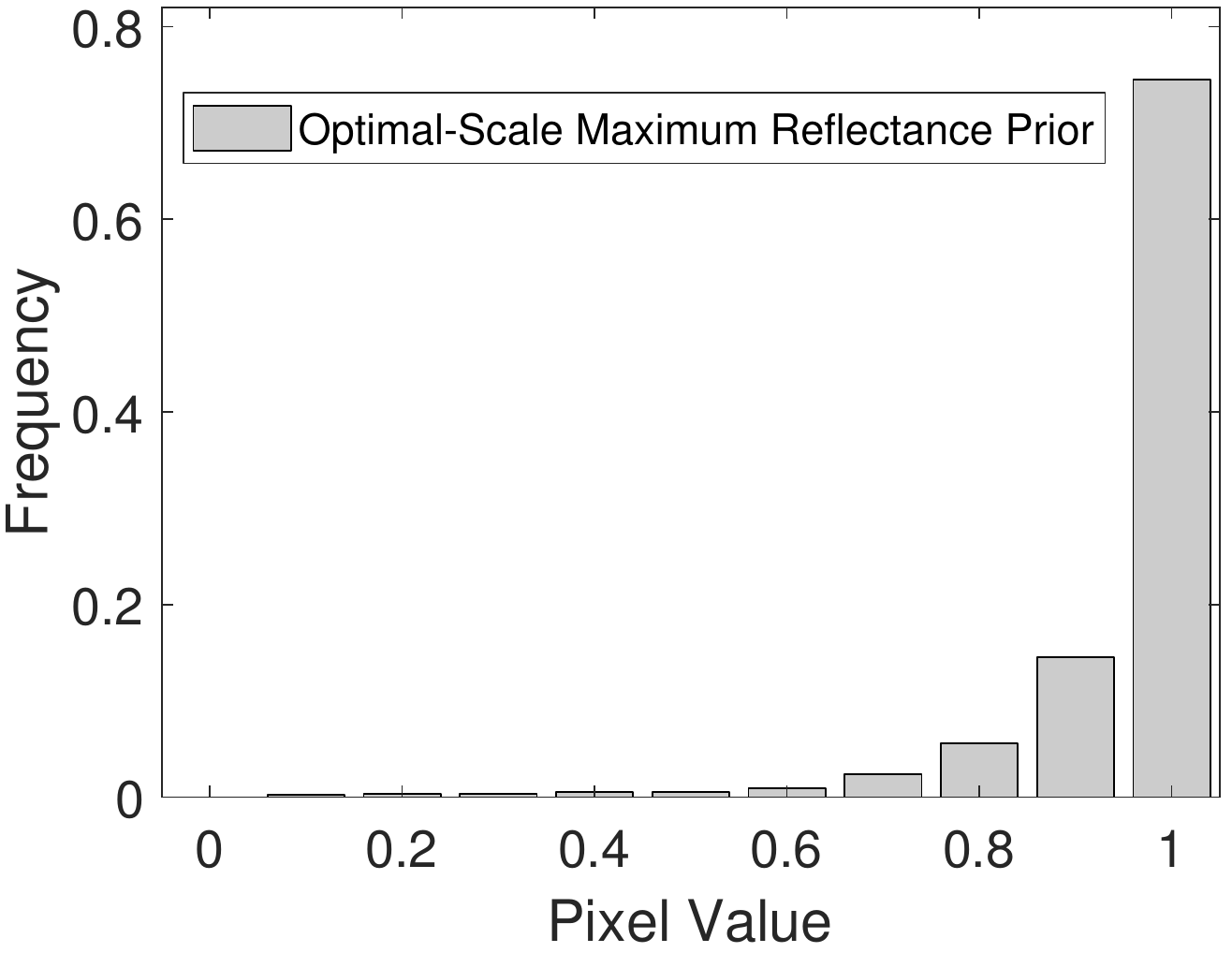}}
\hspace{0.001\linewidth}
\subfloat[]{\label{fig:mmrp_scale_hist}
\includegraphics[width=0.245\linewidth]{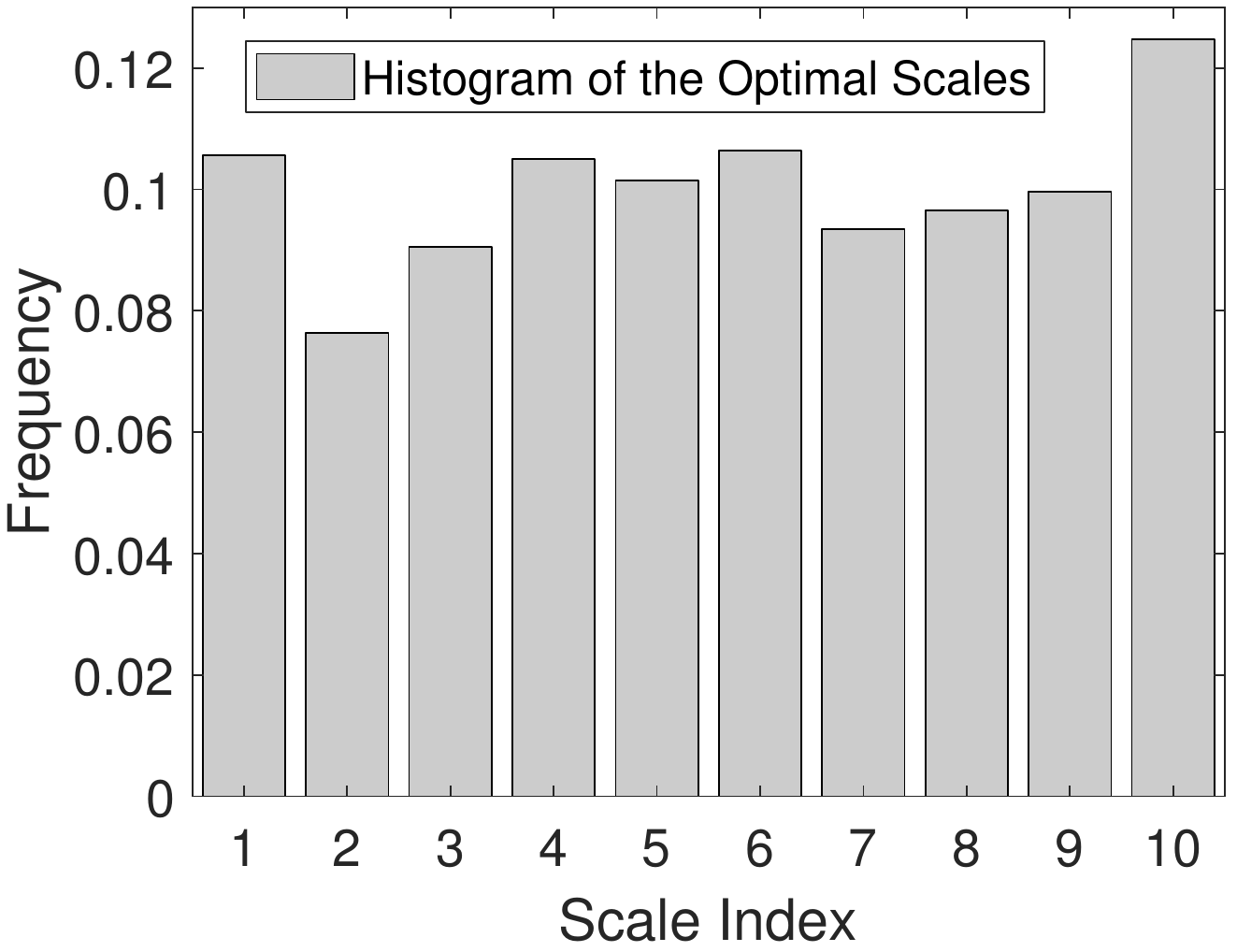}}

\caption{(a)-(d) Histograms of DCP \cite{he2009single}, MRP \cite{zhang2017fast}, the proposed OS-MRP, and the optimal scales in OS-MRP, respectively. 
}
\label{fig:osmrp_stats}
\end{figure*}

\section{OS-MRP for nighttime dehazing}
\label{sec:osfd}

\subsection{Optimal-scale maximum reflectance prior}
\label{subsec:OSMRP}

Following \cite{zhang2017fast}, we define the maximum reflectance $\rmbm{M_s}$ of a daytime clear image patch $\rmbm{R_{{\Omega _s}}}$ centered at $\rmbm{x}$ as:
\begin{equation}
{M_{sc} \left (\rmbm{x} \right) } = \mathop {\max }\limits_{\rmbm{x} \in {\Omega _s}} {R_c}\left( \rmbm{x} \right),c \in \left\{ {r,g,b} \right\},
\label{eq:mr}
\end{equation}
where $s$ is the scale of patch $\Omega _s$ and $\rmbm{M_s} = \left[ {{M_{sr}},{M_{sg}},{M_{sb}}} \right]$. 
In \cite{zhang2017fast}, $M_{sc}$ is assumed to be close to 1 for every local patch, $i.e.$, the \emph{MRP}. However, it may not hold for monochromatic areas such as lawn.

To migrate the issue, we calculate $\rmbm{M_s}$ at multiple scales and determine the optimal scale for each pixel from a probability view. Specifically, we can treat the maximum reflectance as the probability of a surface patch that completely reflects incident light at all frequency range, $i.e.$,
\begin{equation}
\aboveskip
    {P_s}\left (\rmbm{x} \right) = \prod\limits_{c \in \left\{ {r,g,b} \right\}} {{M_{sc}\left (\rmbm{x} \right)}}.
\label{eq:prob}
\belowskip
\end{equation}
For patches containing distinct colors or bright pixels, as shown in the blue rectangles in Figure~\ref{fig:osmrp_demo_real}(a), ${P_s}$ of a small patch is close to 1. For some monochromatic areas, $e.g.$, the left-most green rectangle region, it needs to enlarge the patch size to include more diverse pixels. However, a large patch means a bad localization when estimating the local color cast. To address this issue, we define an optimal scale that is sufficiently large but not necessary to be larger to obtain the highest ${P_s}$. Since ${P_s}$ is a monotonically increasing function of $s$, the optimal scale is defined as:
\begin{equation}
\aboveskip
{s^*}\left (\rmbm{x} \right) = \mathop {\min }\limits_s \left\{ {\mathop {\arg \max }\limits_{\forall s \in \mathcal{S} } {P_s}\left (\rmbm{x} \right)} \right\},
\label{eq:optimalScale}
\belowskip
\end{equation}
where $\mathcal{S}$ is the set of all scales. Figure~\ref{fig:osmrp_demo_real}(b) shows the optimal scale map of (a). The optimal scales are very small for pixels with distinct colors or bright pixels. By contrast, the optimal scale is very large for monochromatic pixels. Based on the above definition, we assume:
\begin{equation}
{M_{{s^*}c}}\left( \rmbm{x} \right) \approx 1,\forall c \in \left\{ {r,g,b} \right\},\forall \rmbm{x} \in {\rm \mathcal{X}},
\label{eq:osmrp}
\end{equation}
where $\mathcal{X}$ is the index set of all pixels. In this paper, we call it \emph{optimal-scale maximum reflectance prior (OS-MRP)}.

To validate this prior, we calculated the histograms of $P_{s^*}$ over 1,000,000 patches sampled from daytime clear images by referring to \cite{he2009single,zhang2017fast}. Ten scales ranging from 7x7 to 43x43 were used when calculating $P_{s^*}$. A fixed scale of 25x25 was used for DCP and MRP. The results are plotted in Figure~\ref{fig:osmrp_stats}. Compared with MRP, more patches have the maximum reflectance in our optimal-scale case, $i.e.$, a higher bar in the last range, and lower bars in the others. Further, the optimal scale is nearly uniformly distributed as shown in Figure~\ref{fig:osmrp_stats}(d), implying that OS-MRP does not obtain unfair advantages from larger patches over MRP.

\subsection{Nighttime dehazing}
\label{subsec:ND-OSMRP}
\subsubsection{Initial multiscale fusion}
\label{subsubsec:msfd}
Since ${P_s}\left (\rmbm{x} \right)$ should be calculated on the clear image, we propose a novel algorithm by first dehazing using multi-scale fusion then refining it using optimal-scale fusion.

\textbf{Color cast correction:}
Following \cite{zhang2017fast}, we also assume ${L}$, $\rmbm{\eta}$, and $t$ are constant at each local patch ${{{\Omega _s}}}$, $i.e.$, ${L_s}$, $\rmbm{\eta_s}$, and $t_s$. Then, we calculate the maximum on ${{{\Omega _s}}}$ as:
\begin{equation}
\aboveskip
\begin{aligned}
 \mathop {\max }\limits_{\rmbm{x} \in {{\Omega _s}}} {I_c}\left( \rmbm{x} \right) &= {\mathop {\max }\limits_{\rmbm{x} \in {{\Omega _s}}} {R_c}\left( \rmbm{x} \right)} {{L_{s}}}{{\eta _{sc}}}{t_s} + {{L_{s}}}{{\eta _{sc}}}\left( {1 - {t_s}} \right) \\
  &= {{L_{s}}}{{\eta _{sc}}},
\label{eq:mrpEta}
\end{aligned}
\belowskip
\end{equation}
We get the last equality using MRP, $i.e.$, {$\mathop {\max }\limits_{\rmbm{x} \in {{\Omega _s}}} {R_c}\left( \rmbm{x} \right) = 1$}. Given the illuminance is the maximum of all channels, $i.e.$,
\begin{equation}
\aboveskip
    \widehat{{L_{s}}} = \mathop {\max }\limits_{c \in \left\{ {r,g,b} \right\}} \mathop {\max }\limits_{\rmbm{x} \in {{\Omega _s}}} {I_c}\left( \rmbm{x} \right),
\label{eq:L}
\belowskip
\end{equation}
we can get the color cast accordingly,
\begin{equation}
\aboveskip
    \widehat{{\eta _{{\rm{sc}}}}} = {{\mathop {\max }\limits_{\rmbm{x} \in {{\Omega _s}}} I_c\left( \rmbm{x} \right)} \mathord{\left/
     {\vphantom {{\mathop {\max }\limits_{x \in {\Omega _s}} I_c\left( x \right)} {\widehat{{L_{s}}}}}} \right.
     \kern-\nulldelimiterspace} {\widehat{{L_{s}}}}}.
\label{eq:eta}
\belowskip
\end{equation}
We then average $\rmbm{\eta_s}$ at each scale to obtain an initial estimate:
\begin{equation}
\aboveskip
    \rmbm{\widehat\eta}  = \frac{1}{{\left| \mathcal{S} \right|}}\sum\limits_{s \in \mathcal{S}} \rmbm{\widehat{{\eta _s}}},
\label{eq:etaFusion}
\belowskip
\end{equation}
where ${\left| \mathcal{S} \right|}$ is the number of scales. We use the fast guided filter \cite{he2015fast} to refine $\rmbm{\eta}$ due to its low computational cost. Then, we remove the color cast according to Eq.~\eqref{eq:imagingModel}:
\begin{equation}
\aboveskip
    \rmbm{I}\left( \rmbm{x} \right) = \rmbm{J}\left( \rmbm{x} \right)t\left( \rmbm{x} \right) + {L}\left( \rmbm{x} \right)\left( {1 - t\left( \rmbm{x} \right)} \right).
\label{eq:colorCorrection}
\belowskip
\end{equation}
Here, we reuse $\rmbm{I}\left( \rmbm{x} \right)$ to denote the color correction result for simplicity. Although we can average $L_s$ to obtain a fusion estimate like $\rmbm{\eta_s}$, the illuminance intensity is less smooth than the color cast due to depth, occlusion, $etc$. Therefore, we re-estimate $L$ using MRP on Eq.~\eqref{eq:colorCorrection} like in Eq.~\eqref{eq:L}.

\textbf{Dehazing:}
Given $L$, it is straightforward to obtain $t$ using the DCP \cite{he2009single} on Eq.~\eqref{eq:colorCorrection}, $i.e.$,
\begin{equation}
\aboveskip
    t = 1 - {{\mathop {\min }\limits_{x \in {{\Omega _t}}} \mathop {\min }\limits_{c \in \left\{ {r,g,b} \right\}} {I_c}\left( x \right)} \mathord{\left/
     {\vphantom {{\mathop {\min }\limits_{x \in {{\Omega _t}}} \mathop {\min }\limits_{c \in \left\{ {r,g,b} \right\}} {I_c}\left( x \right)} {\mathop {\min }\limits_{x \in {{\Omega _t}}} L}}} \right.
     \kern-\nulldelimiterspace} {\mathop {\min }\limits_{x \in {{\Omega _t}}} L}}\left( x \right),
\label{eq:t}
\belowskip
\end{equation}
where ${\Omega _t}$ is the local patch. We refine $t$ using the fast guided filter. Finally, the haze-free image is recovered as:
\begin{equation}
\aboveskip
    \rmbm{J} = {{\left( {\rmbm{I} - L} \right)} \mathord{\left/
     {\vphantom {{\left( {\rmbm{I} - L} \right)} {\max \left( {t,{t_0}} \right) + }}} \right.
     \kern-\nulldelimiterspace} {\max \left( {t,{t_0}} \right) + }}L,
\label{eq:dehazing}
\belowskip
\end{equation}
where $t_0$ is a small threshold for numerical stability.

\subsubsection{OSFD: Optimal-scale fusion-based dehazing}
\label{subsubsec:osfd}
Based on the initial estimate $\rmbm{J}$ and $L$, we normalize $\rmbm{J}$ by $L$ according to Eq.~\eqref{eq:J} and calculate the optimal scale $s^{*}\left ( \rmbm{x} \right)$ at all pixels according to Eq.~\eqref{eq:mr} $\sim$ Eq.~\eqref{eq:optimalScale}. The optimal scale can be regarded as the ``best'' patch size on which to estimate the color cast, where there are sufficient maximum reflectance pixels to be used. Therefore, we propose the following optimal-scale fusion to estimate the color cast:
\begin{equation}
\rmbm{\widehat\eta}  = \sum\limits_{s \in \mathcal{S}} {{\delta _{{s^{*}}s}}\rmbm{\widehat{{\eta _s}}}},
\label{eq:etaFusionOptimal}
\end{equation}
where ${\delta _{{s^*}s}}$ is the Kronecker delta function, $i.e.$, ${\delta _{{s^{*}}s}} = 1$ if $s^{*} = s$, otherwise 0. After obtaining $\rmbm{\eta}$, we remove the color cast and haze according to Eq.~\eqref{eq:colorCorrection} $\sim$ Eq.~\eqref{eq:dehazing}.

\subsection{Computational complexity analysis}
\label{subsec:complexity}
There are two kinds of basic operation in OSFD: the patch-wise \emph{max}, \emph{min}, \emph{sum}, and the pixel-wise \emph{sum}, \emph{multiplication}, \emph{division}. To avoid the dense patch-wise \emph{max} and \emph{min}, we adopt an overlapping sliding window algorithm, reducing the complexity from $\mathcal{O}\left( r^{2}N \right)$ to $\mathcal{O}\left( N \right)$. $r$ is the patch radius. $N$ is the number of pixels. The stride of the sliding window is $r$. For the patch-wise \emph{sum}, we follow \cite{he2015fast} and use the summed-area table algorithm, reducing the complexity from $\mathcal{O}\left( r^{2}N \right)$ to $\mathcal{O}\left( N \right)$. Moreover, compared with the original guided filter \cite{he2010guided}, the computational complexity of the fast guided filter is reduced from $\mathcal{O}\left( N \right)$ to $\mathcal{O}\left( {N \mathord{\left/
 {\vphantom {N {{d^2}}}} \right.
 \kern-\nulldelimiterspace} {{d^2}}} \right)$, where $d$ is the sub-sampling ratio. Therefore, considering that $\rmbm{\eta_s}$ is calculated at all $\left|S\right|$ scales, the total computational complexity of OSFD is $\mathcal{O}\left( \left | \mathcal{S} \right |N \right)$. Further, we down-sample the image $\rmbm{I}$ by a ratio $k_s$ when calculating $\rmbm{\eta_s}$, where ${k_s} \ge 1$ is the ratio of the patch radius between scale $s$ and the lowest scale. Instead of using a larger patch on $\rmbm{I}$, we use a fixed-size patch on the down-sampled image $\rmbm{I_s}$, which reduces the complexity of patch-wise \emph{max} from $\mathcal{O}\left( N \right)$ to $\mathcal{O} \left( {\frac{1}{{k_s^2}}N} \right )$. Although the total computational complexity is still $\mathcal{O}\left( \left | \mathcal{S} \right |N \right)$, it is faster.

\section{A CNN-based Baseline}
\label{sec:dcnn}

\subsection{Network structure}
\label{subsec:network}
In this paper, we devise a simple baseline model named ND-Net based on the deep convolutional neural network. Inspired by the recent success of the encoder-decoder structure in image restoration tasks including daytime dehazing \cite{zhang2019famed}, we also adopt an encoder-decoder structure which consists of a MobileNet-v2 backbone as the encoder and a fully convolutional decoder. Considering that nighttime image dehazing tries to learn the inverse mapping of Eq.~\eqref{eq:imagingModel}, which is less complex than large vision tasks, we use the MobileNet-v2 as the encoder backbone due to its computational efficiency and lightweight parameters. Since this part of the work is to provide a feasible learning-based solution and validate the usage of the proposed synthetic benchmark, we leave it as the future work to explore other backbones and devise specific dehazing blocks.

The decoder has five convolutional blocks and a convolutional prediction layer. Each convolutional block has two branches with the similar structure of the first block of each stage of the ResNet. One branch is for residual projection and has a convolutional layer followed by a Batch Normalization layer and a ReLU layer. The other branch has a bottleneck structure that consists of an 1*1 convolutional layer for feature dimension reduction, a 3*3 convolutional layer, and an 1*1 convolutional layer for feature dimension recovery. Each convolutional layer is followed by a Batch Normalization layer and a ReLU layer. After each block, we use bilinear interpolation to increase the feature resolution to its counterpart from each stage in the encoder and add them together via an element-wise sum. In other words, we add skip connections between corresponding encoder and decoder blocks to form a U-Net structure and leverage the encoded features at different levels and scales for decoding. The final prediction layer is an 1*1 convolutional layer without Batch Normalization and ReLU layers. We leverage the residual learning idea by adding a skip connection between the input hazy image and the prediction output via element-wise sum, making the encoder-decoder network to learn the haze residual rather than a clear image, which is more effective and easier to train.

\subsection{Training objectives}
\label{subsec:loss}
We use the Mean Square Error (MSE) loss and the perceptual loss based on the pretrained VGG network \cite{simonyan2015very,johnson2016perceptual}, $i.e.$,
\begin{equation}
L_{mse} = {\left\| {\rmbm{J} - \rmbm{J_{GT}}} \right\|_2},
\label{eq:mseloss}
\end{equation}
\begin{equation}
L_{perceptual} = \sum\limits_{l \in \left\{ {4,9,16,23} \right\}} {{{\left\| {vg{g_l}\left( \rmbm{J} \right) - vg{g_l}\left( {\rmbm{J_{GT}}} \right)} \right\|}_1}},
\label{eq:vggloss}
\end{equation}
where $\rmbm{J_{GT}}$ is the ground truth haze-free image, ${vg{g_l}\left(  \cdot  \right)}$ denotes the feature map from the $l^{th}$ layer of the VGG network, ${\left\|  \cdot  \right\|_2}$ and ${\left\|  \cdot  \right\|_1}$ denote the L2 and L1 norms, respectively. The final training objective is a linear combination of both losses, $i.e.$,
\begin{equation}
L = L_{mse} + \lambda L_{perceptual},
\label{eq:finalloss}
\end{equation}
where $\lambda$ is a loss weight.

\section{Experiments}
\label{sec:experiment}

\subsection{Experimental settings}
\label{subsec:experimentsettings}
\textbf{Datasets and evaluation metrics:}
We used the synthetic datasets in Section~\ref{subsec:benchmark} to benchmark SOTA methods and our OSFD. We also used the 150 real-world nighttime hazy images from \cite{zhang2017fast,li2015nighttime} for subjective evaluation, denoted NHRW. We collected 1,500 real-world daytime clear images for color removal evaluation, denoted DCRW. The statistics of these datasets are listed in Table~\ref{tab:datasetStats}. PSNR, SSIM \cite{wang2004image}, and CIEDE2000 \cite{sharma2005ciede2000} are used as the evaluation metrics.

\begin{table}[htbp]
  \centering
  \caption{Statistics of the benchmark datasets. ND and CR refer to the nighttime dehazing task and color removal task. }
    \begin{tabular}{lccccc}
    \toprule
    Dataset & Haze Density  & Number & Synthetic    & Tasks\\
    \midrule
    NHC-L & Light & 2,750  & \checkmark     & ND+CR \\
    NHC-M & Medium & 2,750  & \checkmark     & ND \\
    NHC-D & Dense & 2,750  & \checkmark     & ND \\
    NHM   & All   & 350   & \checkmark     & ND \\
    NHR   & All   & 8,970  & \checkmark     & ND \\
    NHRW  & All   & 150   & \text{\sffamily x}     & ND \\
    DCRW  & \text{\sffamily x}    & 1500  & \text{\sffamily x}     & CR \\
    \bottomrule
    \end{tabular}%
  \label{tab:datasetStats}%
\end{table}%

\textbf{Implementation details:}
We used 10 scales in OS-MRP, where the size of $\Omega_s$ ranged from 7x7 to 43x43. The size of $\Omega_t$ was set to 15x15. $\beta_l$ was set to 1. $\beta_t$ was set to 0.005, 0.1, and 0.2. $t_0$ was set to 0.1. The loss weight $\lambda$ was set to 0.01. OSFD was implemented in C++. 3R was implemented in MATLAB. ND-Net was implemented with PyTorch and trained on a single TITAN Tesla V100 GPU. We used the NHR dataset to train ND-Net since it contains diverse images. NHR was split into two disjoint parts, $i.e.$, 8073 images as the training set, and 897 images as the validation set. We also evaluated the model on other benchmark datasets listed in Table~\ref{tab:datasetStats}. We will release the source codes and datasets for reproducibility.

\subsection{Main results}
\label{subsec:mainResults}

\begin{table}[htbp]
  \centering
  \caption{Dehazing results on the NHC dataset.}
    \begin{tabular}{p{2.6cm}p{1.1cm}<{\centering}p{1.1cm}<{\centering}p{1.8cm}<{\centering}}
    \toprule
      Method    & PSNR\,($\uparrow$)  & SSIM\,($\uparrow$)  & CIEDE2000\,($\downarrow$) \\
    \midrule
    & \multicolumn{3}{c}{NHC-L}\\
    \hline
    NDIM \cite{zhang2014nighttime}  & 11.12 & 0.2867 & 21.77 \\
    GS \cite{li2015nighttime} & 18.84 & 0.5537 & 10.12 \\
    MRPF \cite{zhang2017fast} & 19.17 & 0.5831 & 9.42 \\
    MRP \cite{zhang2017fast} & 23.02 & 0.6855 & 8.12 \\
    \textbf{OSFD} & \textbf{23.10} & \textbf{0.7376} & \textbf{7.48} \\
    \textbf{ND-Net} & \underline{\textbf{26.12}} & \underline{\textbf{0.8519}} & \underline{\textbf{6.16}} \\
    \hline
    & \multicolumn{3}{c}{NHC-M}\\
    \hline
    NDIM \cite{zhang2014nighttime}  & 10.93 & 0.2959 & 22.31 \\
    GS \cite{li2015nighttime} & 15.88 & 0.4654 & 12.81 \\
    MRPF \cite{zhang2017fast} & 16.40 & 0.5093 & 11.71 \\
    MRP \cite{zhang2017fast} & 20.61 & 0.6238 & 9.12 \\
    \textbf{OSFD} & \textbf{21.15} & \textbf{0.6782} & \textbf{8.56} \\
    \textbf{ND-Net} & \underline{\textbf{22.72}} & \underline{\textbf{0.7899}} & \underline{\textbf{7.31}} \\
    \hline
    & \multicolumn{3}{c}{NHC-D}\\
    \hline
    NDIM \cite{zhang2014nighttime}  & 10.66 & 0.3077 & 23.25 \\
    GS \cite{li2015nighttime} & 12.70 & 0.3690 & 17.72 \\
    MRPF \cite{zhang2017fast} & 13.48 & 0.4289 & 15.71 \\
    MRP \cite{zhang2017fast} & 17.62 & 0.5483 & 11.24 \\
    \textbf{OSFD} & \textbf{18.41} & \textbf{0.6002} & \textbf{10.66} \\
    \textbf{ND-Net} & \underline{\textbf{18.90}} & \underline{\textbf{0.7010}} & \underline{\textbf{9.58}} \\
    \bottomrule
    \end{tabular}%
  \label{tab:NHC}%
\end{table}%

\begin{table*}[htbp]
  \centering
  \caption{Dehazing results on the NHM and NHR datasets.}
    \begin{tabular}{lccc|ccc}
    \toprule
    & \multicolumn{3}{c}{NHM} & \multicolumn{3}{c}{NHR}\\
    \hline
      Method    & PSNR\,($\uparrow$)  & SSIM\,($\uparrow$)  & CIEDE2000\,($\downarrow$) & PSNR\,($\uparrow$)  & SSIM\,($\uparrow$)  & CIEDE2000\,($\downarrow$) \\
    \midrule
    NDIM \cite{zhang2014nighttime}  & 14.58 & 0.5630 & 19.23 & 14.31 & 0.5256 & 18.15 \\
    GS \cite{li2015nighttime} & 16.84 & 0.6932 & 15.84 & 17.32 & 0.6285 & 12.32 \\
    MRPF \cite{zhang2017fast} & 13.85 & 0.6056 & 19.20 & 16.95 & 0.6674 & 12.32 \\
    MRP \cite{zhang2017fast} & 17.74 & 0.7105 & 15.23 & 19.93 & 0.7772 & 10.01 \\
    \textbf{OSFD} & \textbf{19.75} & \textbf{0.7649} & \textbf{12.23} & \textbf{21.32} & \textbf{0.8035} & \textbf{8.67} \\
    \textbf{ND-Net} & \underline{\textbf{21.55}} & \underline{\textbf{0.9074}} & \underline{\textbf{9.11}} & \underline{\textbf{28.74}} & \underline{\textbf{0.9465}} & \underline{\textbf{4.02}} \\
    \bottomrule
    \end{tabular}%
  \label{tab:NHMR}%
\end{table*}%

\begin{figure*}
  \centering
  \includegraphics[width=1\linewidth]{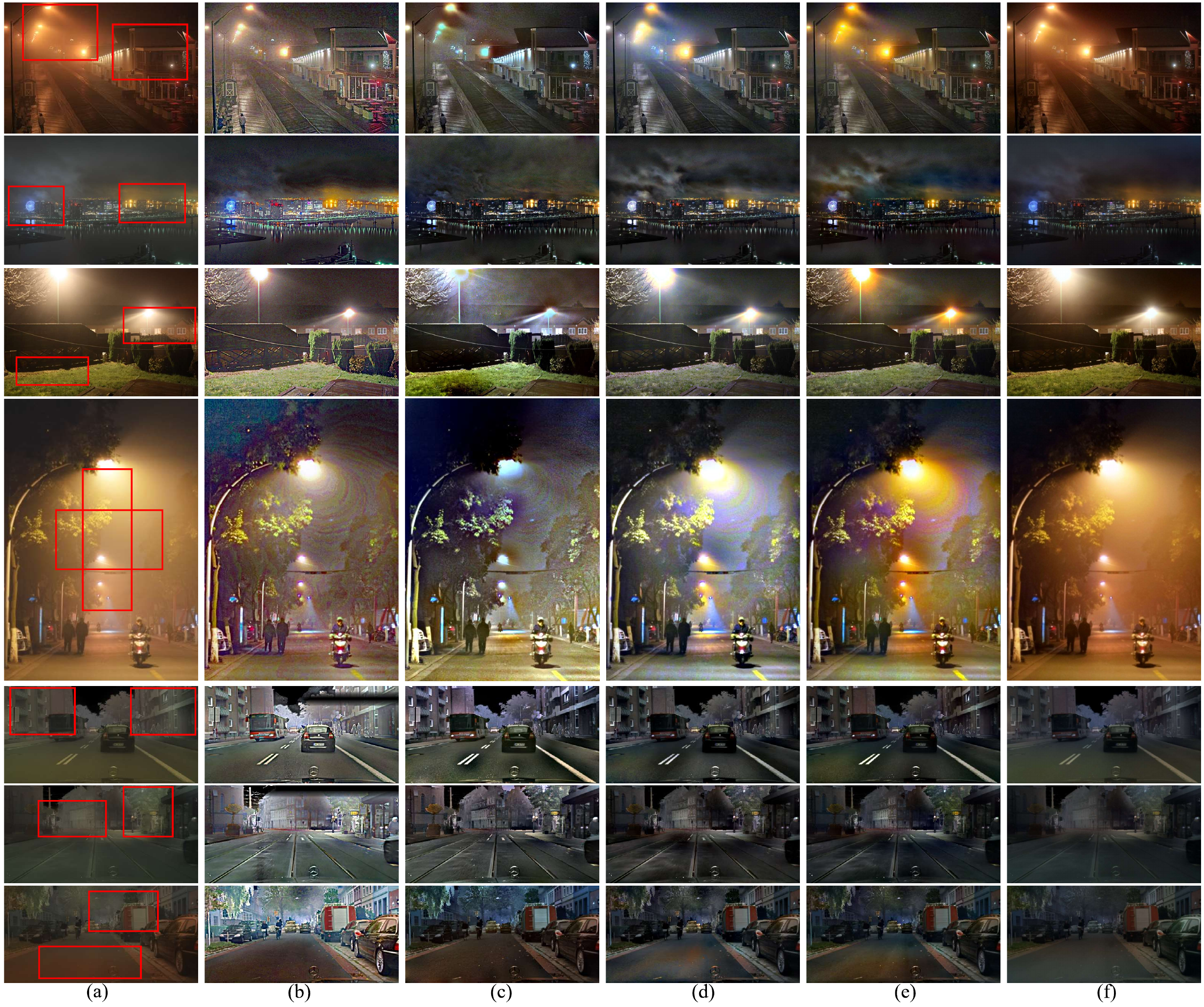}
  \captionof{figure}{(a) Nighttime hazy images. (b) NDIM \cite{zhang2014nighttime}. (c) GS \cite{li2015nighttime}. (d) MRP \cite{zhang2017fast}. (e) Our OSFD. (f) Our ND-Net.}
  \label{fig:subjectiveEvalExp}
\end{figure*}

\textbf{Evaluation on nighttime dehazing:}
We compared the proposed methods with nighttime dehazing methods including NDIM \cite{zhang2014nighttime}, GS \cite{li2015nighttime}, MRP and MRPF \cite{zhang2017fast} according to objective metrics and subjective inspection. The quantitative results on NHC-L, NHC-M, NHC-D, NHM, and NHR are listed in Table~\ref{tab:NHC} and~\ref{tab:NHMR}. As can be seen, with increasing haze density, it becomes difficult to remove the color cast and haze, and the performance of all the methods degrades. NDIM achieves the worst scores for all the metrics. From Figure~\ref{fig:banner}(b) and Figure~\ref{fig:subjectiveEvalExp}(b), we can see that NDIM tends to generate a bright dehazed image but with color artifacts and amplified noise. GS's performance is much better than NDIM, $e.g.$, less haze left in the results. As can be seen from Figure~\ref{fig:banner}(c) and Figure~\ref{fig:subjectiveEvalExp}(c), GS removes the glows and generates clear images. However, it also produces artifacts around light sources and distinct edges with 
sharp intensity discontinuity.

The proposed OSFD achieves the best scores among all the prior-based methods, demonstrating that it can remove haze, recover details, and preserve local structures efficiently (see red rectangle regions). It outperforms MRP by a margin of about 0.05 SSIM score on NHC-L, NHC-M, and NHC-D, validating the superiority of OS-MRP over MRP. As can be seen from Figure~\ref{fig:banner}(d)-(e) and Figure~\ref{fig:subjectiveEvalExp}(d)-(e), MRP fails with the monochromatic road, wall, light source, and lawn regions, leading to whitish results due to the inaccurate color cast estimate biased towards the intrinsic colors of images. On the contrary, the proposed OS-MRP works well, and leads to more visually realistic results. As for the proposed CNN-based baseline ND-Net, it achieved the best scores in terms of all metrics. Note that ND-Net was trained on the NHR dataset and cross-validated on the NHC ones. These results in Table~\ref{tab:NHC} confirms that the ND-Net trained on NHR has a good generalization ability. The dehazed results by ND-Net in Figure~\ref{fig:banner}(f) and Figure~\ref{fig:subjectiveEvalExp}(f) are more visual pleasing than others, $e.g.$, less color, illumination, and noise artifacts. Nevertheless, there is still some residual haze in the results, especially in the light source regions, which can be addressed in the future work by taking the glow effect into account. Results on the NHM and NHR datasets are summarized in Table~\ref{tab:NHMR}. We present more visual results in the supplementary materials.

\begin{table}[htbp]
  \centering
  \caption{Color cast removal results on the NHC-L and DCRW datasets.}
    \begin{tabular}{lccc}
    \toprule
      Method    & PSNR\,($\uparrow$)  & SSIM\,($\uparrow$)  & CIEDE2000\,($\downarrow$) \\
    \midrule
    & \multicolumn{3}{c}{NHC-L}\\
    \hline
    GS \cite{li2015nighttime} & 27.92 & 0.6454 & 5.735 \\
    MRP \cite{zhang2017fast} & 30.80 & 0.7525 & 5.100 \\
    \textbf{OSFD} & \textbf{30.91} & \textbf{0.7586} & \textbf{4.879} \\
    \hline
    & \multicolumn{3}{c}{DCRW}\\
    \hline
    GS \cite{li2015nighttime} & 12.56 & 0.6313 & 19.02 \\
    MRP \cite{zhang2017fast} & 20.29 & 0.7923 & 10.18 \\
    \textbf{OSFD} & \textbf{22.89} & \textbf{0.8711} & \textbf{7.659} \\
    \bottomrule
    \end{tabular}%
  \label{tab:CR}%
\end{table}%

\textbf{Evaluation on color cast removal:}
We also evaluated the performance of GS, MRP, and OSFD on color cast removal. Note that for a clear image without a color cast, it is expected to estimate a white color cast and not change the original colors in the image. Therefore, we test the above methods on the DCRW dataset to see how well they retain the colors. The results are summarized in Table~\ref{tab:CR}. OSFD achieves the best performance on both datasets. It benefits from the proposed OS-MRP, which can adapt to the diverse local statistics in natural images and select the optimal scale to estimate the color cast (see the road, lawn, wall, and light source regions in Figure~\ref{fig:banner} and Figure~\ref{fig:subjectiveEvalExp}). Further, it is noteworthy that all these methods employ the DCP to estimate the transmission and remove haze. As noted above, this prior fails when there is a color cast. Therefore, better color removal performance matters for the subsequent dehazing stage.


\subsection{Running time comparison}
\label{subsec:runningtime}
We evaluated the run-time of the different methods \cite{zhang2014nighttime, li2015nighttime, zhang2017fast} on a PC with an Intel CORE i7 CPU and 16 Gb memory and ND-Net on a Tesla V100 GPU. The codes are from the original authors. NDIM \cite{zhang2014nighttime}, GS \cite{li2015nighttime}, MRPF \cite{zhang2017fast}, MRP \cite{zhang2017fast}, our OSFD and ND-Net process a 512x512 image in 5.63s, 22.52s, 0.236s, 1.769s, 0.576s, and 0.0074s, respectively. Although OSFD calculates $\rmbm{\eta}$ at 10 scales, it is 3x faster than MRP and 10x faster than NDIM and GS. The efficiency of OSFD arises from its $\mathcal{O}\left( \left | \mathcal{S} \right |N \right)$ complexity and the acceleration tricks used in Section~\ref{subsec:complexity}. Since the basic operations in OSFD are paralleled, it promises to be faster using GPU acceleration. The proposed ND-Net is the fastest one and runs at 135 frames per second (FPS), benefiting from its lightweight backbone and GPU acceleration. It is promising to be used for real-time applications.

\subsection{Limitations and discussion}
\label{subsec:limitations}
Although the synthetic nighttime hazy images produced by 3R are visually realistic, they may contain artifacts due to inaccurate depth, i.e., the car boundaries in Figure~\ref{fig:syntheticimamgesdemo}. This can be addressed by using effective depth filtering/completion methods. Further, the light sources are set to isotropic point lights, which can be changed to other forms of light sources, $e.g.$, directional light, and volume light. Also, we can further add the glow images of virtual light sources into the final hazy image according to Eq. (3) in \cite{li2015nighttime} to simulate the glow effect. OSFD may generate results with color artifacts and residual haze due to the following limitations. 1) The statistical prior may not work in large monochromatic areas or dense haze regions with different color casts. 2) The optimal scale may be inaccurate if the initial dehazing stage fails. 3) Since dehazing is disentangled from color cast removal, it may be affected by the residual color cast. For example, see the trees and light sources in the first and fourth rows and the bluish distant areas in the last row in Figure~\ref{fig:subjectiveEvalExp}.

Encouraged by recent progress in daytime dehazing, we are optimistic that deep models have the potential to address these limitations as evidenced by the proposed simple baseline ND-Net. There is still a large room for future studies. 3R generates a large amount of visually realistic training images with intermediate ground truth labels including low-light images, color cast, and illuminance, which can provide auxiliary supervision to train a better model. For example, devising disentangled generative models based on the explicit imaging model and intermediate labels is also very promising.

\section{Conclusion}
We introduce a novel synthetic method called 3R, which leverages both scene geometry and real-world light colors to generate realistic nighttime hazy images. Based on 3R, we construct a new dataset to benchmark state-of-the-art nighttime dehazing methods concerning haze removal, color cast correction, and runtime. We also propose an optimal-scale maximum reflectance prior, which can adapt to the local statistics at varying scales in natural images. Based on this, we disentangle the color correction from haze removal and devise a computationally efficient and effective dehazing method. Besides, we devise a simple but effective CNN-based baseline model, which shows good dehazing and generalization ability. Extensive experiments demonstrate their superiority over SOTA methods in terms of both image quality and runtime.

\section*{Acknowledgments}
\noindent{This work was supported by the National Natural Science Foundation of China (NSFC) under Grants 61806062, 61872327, 61873077, U19B2038, 61620106009, and Australian Research Council Projects under Grant FL-170100117.}

\section{Nighttime Dehazing with a Synthetic Benchmark: Supplementary Material}

\subsection{More synthetic results generated by the proposed 3R}
In this part, we present more visual results synthesized by the proposed 3R (see Section~3 in the paper for more details.). First, we compare the proposed 3R with the synthetic method in \cite{zhang2017fast} in Section~\ref{subsec:syncom}. Then, we present some controllable synthetic results by changing the hyper-parameters in 3R, $e.g.$, $\beta_t$ for haze density, $\eta$ for light colors, and $\beta_l$ for illuminance intensity in Section~\ref{subsec:controllable}. Finally, we also present some synthetic results on the Virtual KITTI dataset \cite{gaidon2016virtual} in Section~\ref{subsec:virtualKitti}.

\subsubsection{Visual comparison with the synthetic method in \cite{zhang2017fast}}
\label{subsec:syncom}
\begin{figure*}[ht]
\centering
\includegraphics[width=1\linewidth]{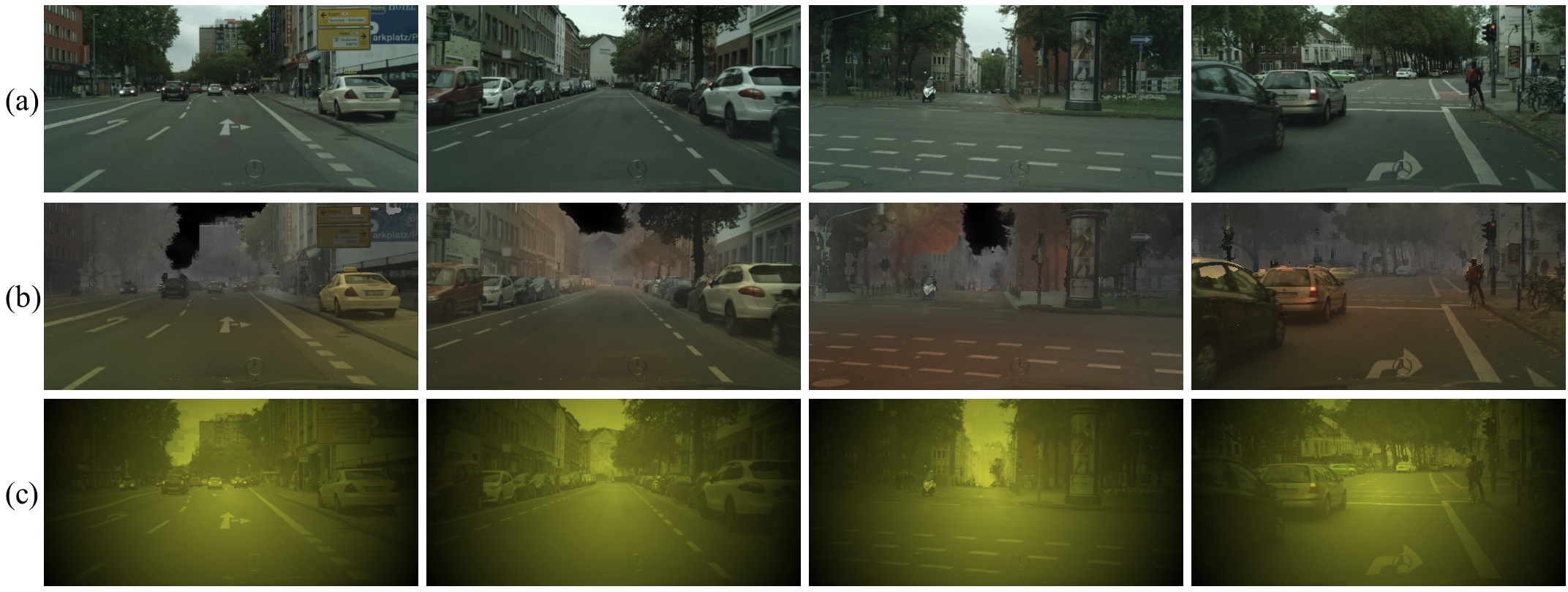}
\caption{(a) Daytime clear images from the Cityscapes dataset \cite{cordts2016cityscapes}. (b) Synthetic results generated by the proposed 3R (see Section~3 in the paper. (c) Results generated by the method in \cite{zhang2017fast}.}
\label{fig:syn_comp1}
\end{figure*}

\begin{figure*}[ht]
\centering
\includegraphics[width=1\linewidth]{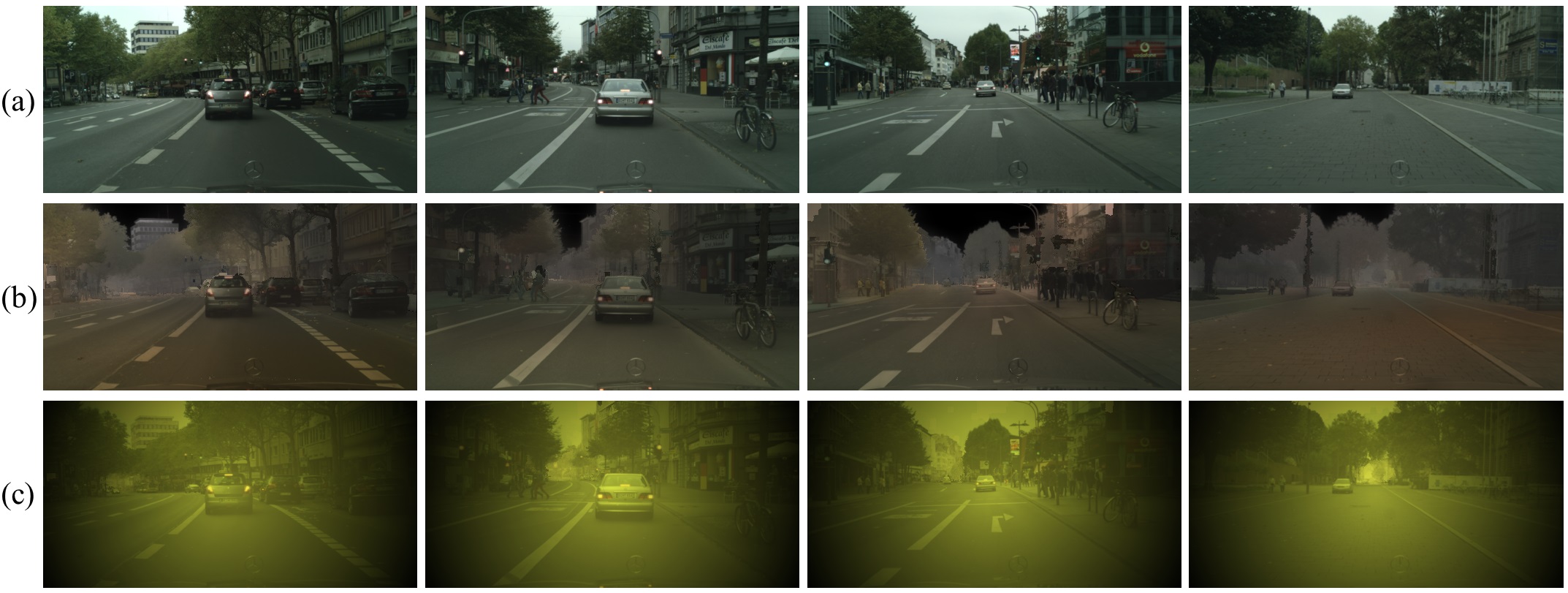}
\caption{(a) Daytime clear images from the Cityscapes dataset \cite{cordts2016cityscapes}. (b) Synthetic results generated by the proposed 3R (see Section~3 in the paper.). (c) Results generated by the method in \cite{zhang2017fast}.}
\label{fig:syn_comp2}
\end{figure*}

Figure~\ref{fig:syn_comp1} and \ref{fig:syn_comp2} show the synthetic nighttime hazy images generated by the proposed 3R and the method in \cite{zhang2017fast}. As can be seen, the overall light colors are uniform, $i.e.$, yellow, in Figure~\ref{fig:syn_comp1}(c) and \ref{fig:syn_comp2}(c). By contrast, the light colors in our 3R's results are more realistic, $i.e.$, changed from light yellow to warm red. Meanwhile, the color cast at different areas within an image may be different since we randomly sampled light colors from the prior distribution (see Section~3.1 in the paper). Consequently, it is more challenging to deal with the non-uniform color cast.

Further, the results generated by \cite{zhang2017fast} have a strong vignetting effect since the method renders the illuminance from a single central light source and the illuminance intensity is only determined by the light path distance. By contrast, on the one hand, we consider the scene geometry when calculating the illuminance, $i.e.$, the illuminance intensity is calculated from the surface normal vector, the incident light direction, and the light path distance according to Lambert's cosine law \cite{basri2003lambertian} (see Section~3.2 in the paper for more details.). On the other hand, we place many virtual light sources along the roadsides, which are 5 meters high and spaced every 30 meters. Consequently, the synthetic illuminance by 3R looks more realistic.

\begin{figure*}
\centering
\includegraphics[width=1\linewidth]{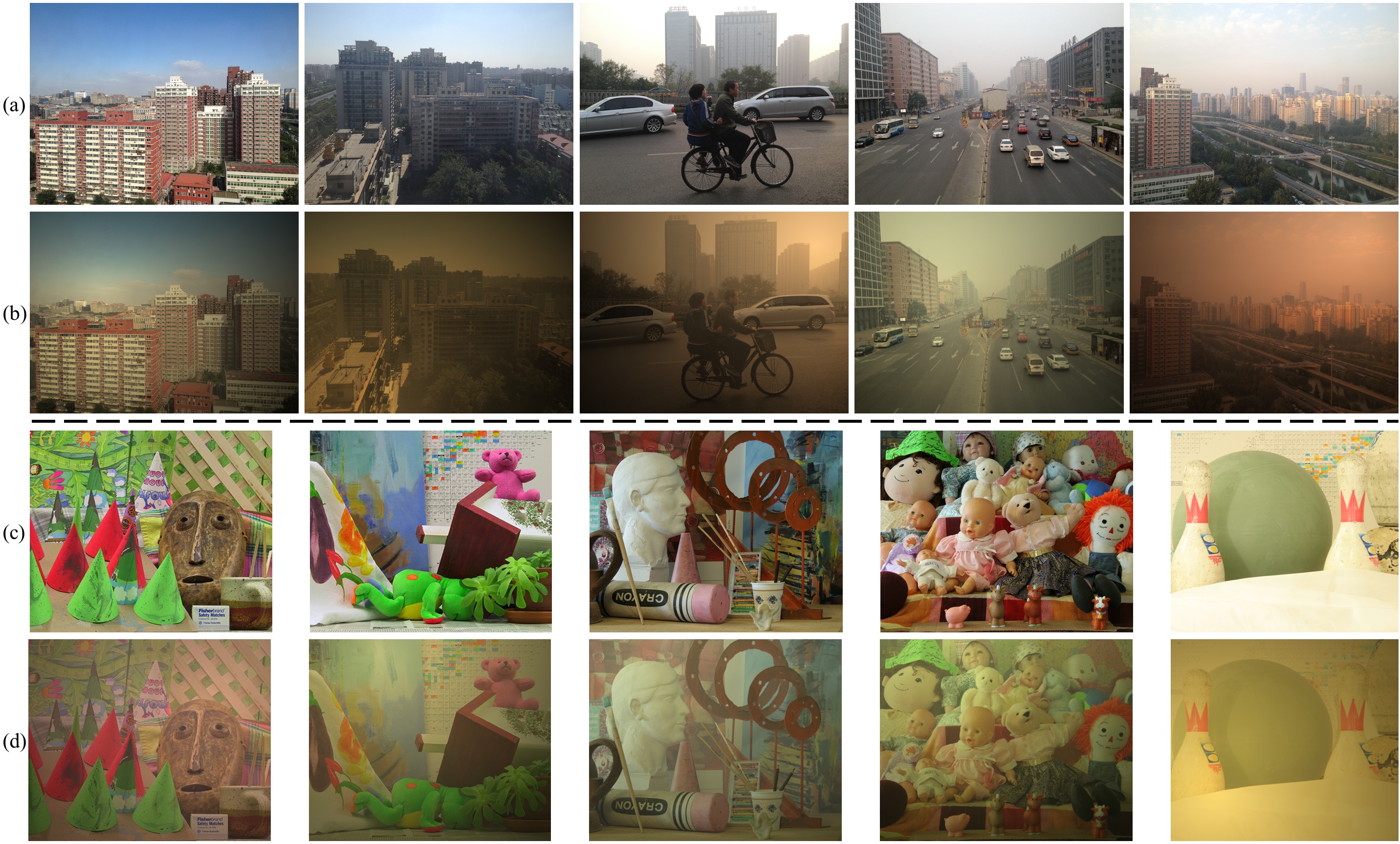}
\caption{(a) Daytime clear images from the RESIDE dataset \cite{li2018benchmarking}. (b) The synthetic nighttime hazy images generated by the method in \cite{zhang2017fast}. The light colors are sampled from the priori distribution in Section~3.1. (c) Daytime clear images from the Middlebury dataset \cite{scharstein2003high}. (d) The synthetic nighttime hazy images generated by the method in \cite{zhang2017fast}. The light colors are sampled from the priori distribution in Section~3.1. }
\label{fig:syn_middleburyReside}
\end{figure*}

In \cite{zhang2017fast}, they synthesized the nighttime hazy images on the Middlebury dataset \cite{scharstein2003high}. However, it is not trivial to use 3R on this dataset since the images are all captured in an indoor environment and there are no semantic labels available. Alternatively, we followed \cite{zhang2017fast} but replaced the constant yellow light color with the randomly sampled real-world light colors described in Section 3.1. Similarly, we also synthesized nighttime hazy images on the RESIDE dataset \cite{li2018benchmarking}. The results in Figure~\ref{fig:syn_middleburyReside} (b) and (d) show more diversity in illuminance colors than Figure~\ref{fig:syn_comp1}(c) and Figure~\ref{fig:syn_comp2}(c). These two synthetic datasets are denoted NHM and NHR, respectively.

\begin{figure*}[ht]
\centering
\includegraphics[width=1\linewidth]{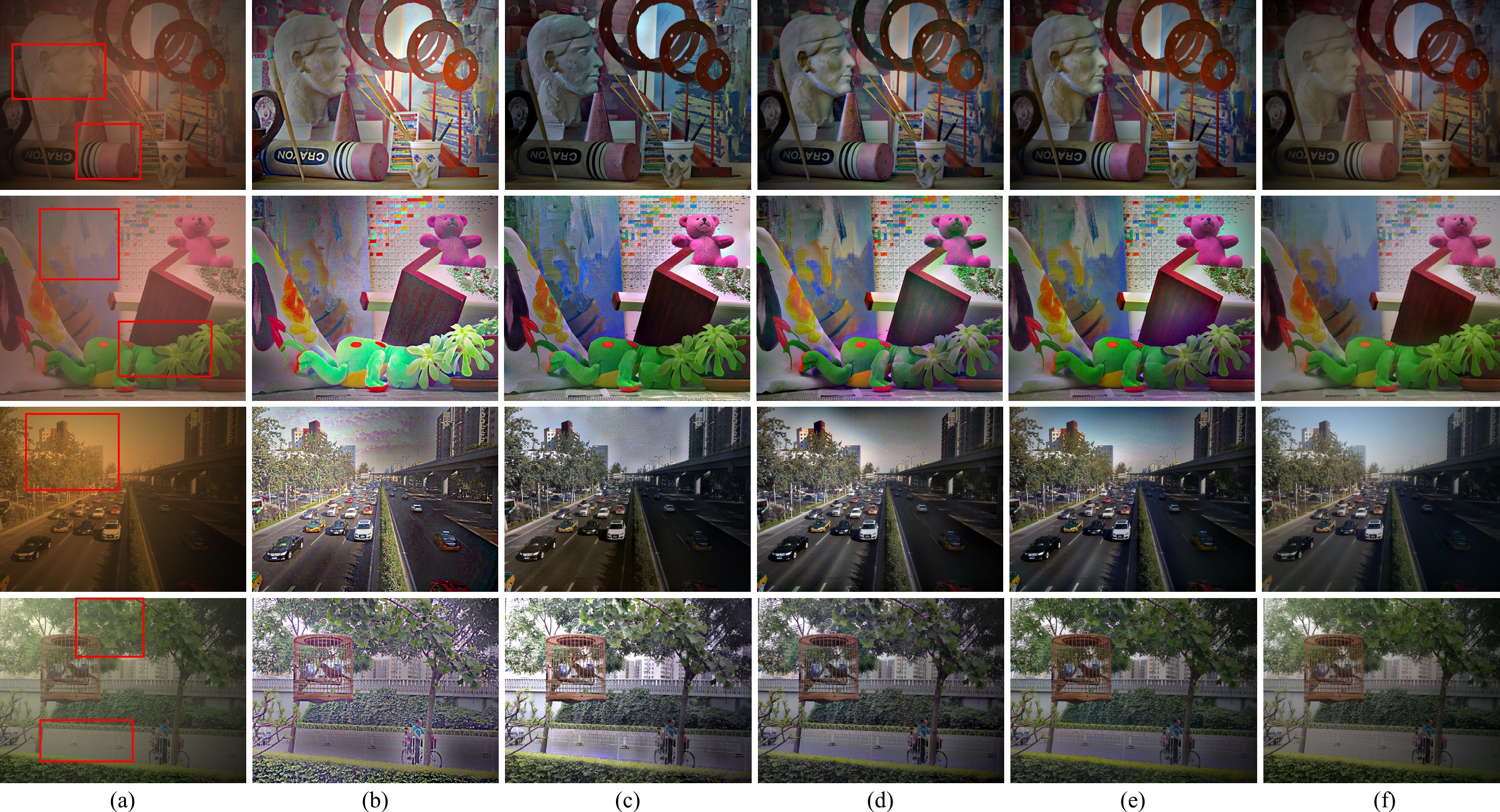}
\caption{(a) Nighttime hazy images from NHM and NHR datasets. (b) NDIM \cite{zhang2014nighttime}. (c) GS \cite{li2015nighttime}. (d) MRP \cite{zhang2017fast}. (e) Our OSFD (see Section~4.2 in the paper). (f) Our ND-Net (see Section~5 in the paper).}
\label{fig:sub_middleburyReside}
\end{figure*}

Some visual results using different methods on NHM and NHR are presented in Figure~\ref{fig:sub_middleburyReside}. Since images in NHM have a shallow depth of field (DOF) in the indoor environment and colorful objects, the methods \cite{zhang2014nighttime, li2015nighttime,zhang2017fast} and our OSFD tend to generate over-dehazed results, resulting in more saturated colors (see the white plaster and the canvas in the first two rows in Figure~\ref{fig:sub_middleburyReside}). Nevertheless, the proposed OSFD achieves better results than others. For example, MRP \cite{zhang2017fast} generates whitish results with color artifacts (see the areas enclosed by the red rectangles.). On the NHR dataset, our OSFD can efficiently remove the haze while retaining the colors as shown in the last two rows in Figure~\ref{fig:sub_middleburyReside}. Generally, the proposed ND-Net achieves the best performance on both NHM and NHR datasets, which efficiently correct the color cast and remove haze. 

\subsubsection{Controllable synthetic results generated by 3R}
\label{subsec:controllable}
In our 3R method, some hyper-parameters such as $\beta_t$, $\eta$, $\beta_l$, can be changed to generate controllable synthetic nighttime hazy images, for example, light haze or dense haze, low illuminance or high illuminance. It could be very useful for training deep neural networks with better generalizability and disentangling the degradation factors in nighttime (hazy) images.

\begin{figure*}[ht]
\centering
\includegraphics[width=1\linewidth]{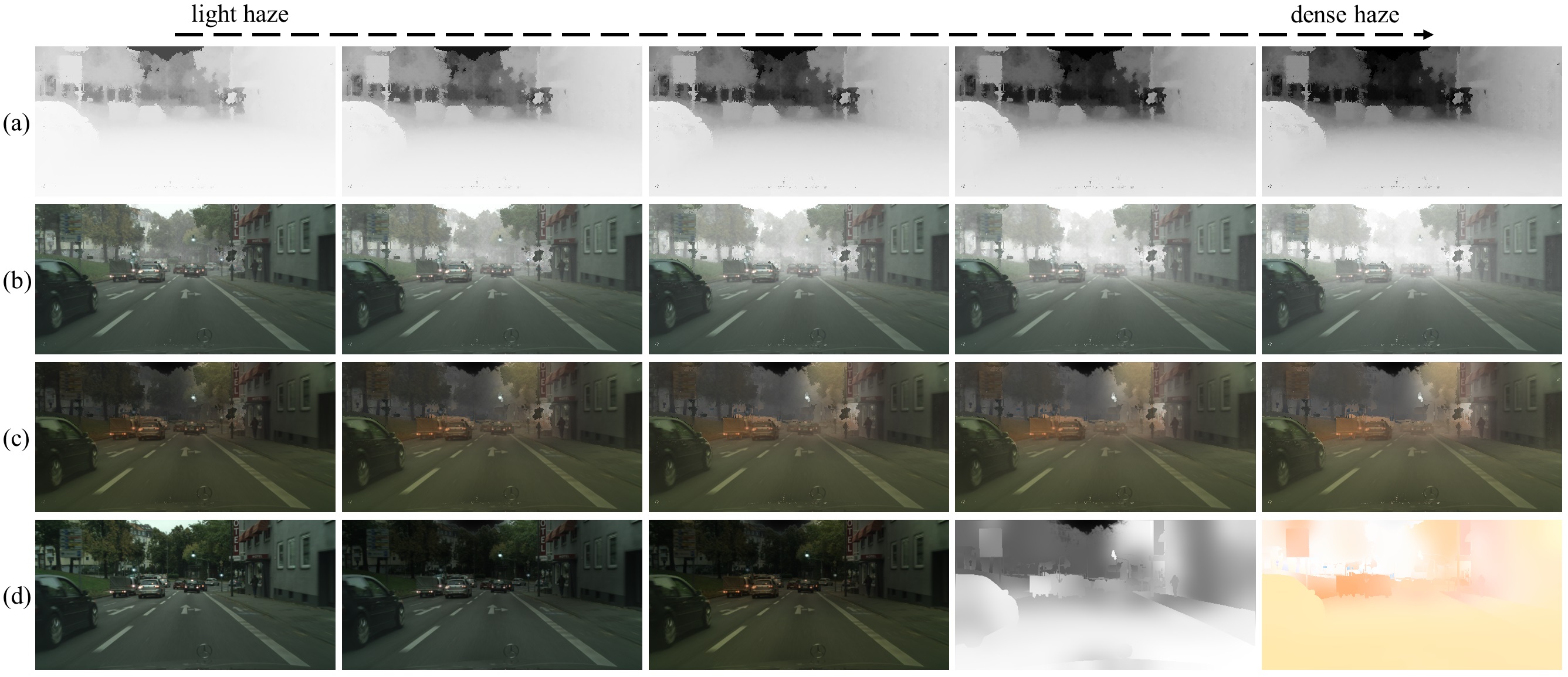}
\caption{(a) The transmission maps at different haze density levels, $i.e.$, $\beta_t = 0.005,0.01,...,0.025$. (b) The synthetic daytime hazy images generated by 3R. (c) The synthetic nighttime hazy images generated by 3R. (d) From left to right, the original daytime clear image, the synthetic nighttime images without/with color cast, the illuminance intensity map, and the color cast map.}
\label{fig:syn_t}
\end{figure*}

\textbf{Haze Density}
As shown in Figure~\ref{fig:syn_t}, with the increasing of $\beta_t$ from 0.005 to 0.025, the transmission maps (Figure~\ref{fig:syn_t}(a)) become darker, and the hazy images (Figure~\ref{fig:syn_t}(b) and (c)) contain more and more haze. It is noteworthy that we kept the illuminance intensity and light colors (Figure~\ref{fig:syn_t}(d)) fixed in this experiment.

\begin{figure*}[ht]
\centering
\includegraphics[width=1\linewidth]{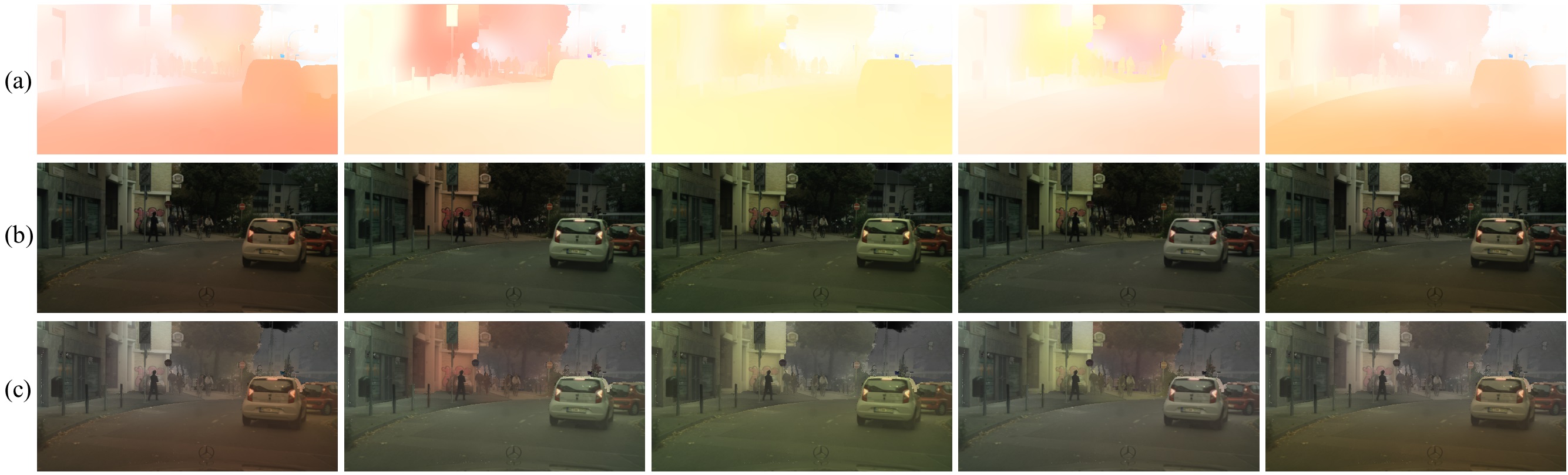}
\caption{(a) Color cast maps by randomly sampling light colors (see Section~3.1 in the paper.). (b) The synthetic nighttime images with color cast generated by 3R. (c) The synthetic nighttime hazy images generated by 3R.}
\label{fig:syn_eta}
\end{figure*}

\textbf{Light Colors}
As shown in Figure~\ref{fig:syn_eta}, we sampled different light colors from the prior distribution (see Section~3.1 in the paper.), resulting in diverse color casts (Figure~\ref{fig:syn_eta}(a)). Accordingly, the synthetic nighttime (hazy) images have more diversity in light colors (Figure~\ref{fig:syn_eta}(b) and (c)). We kept the illuminance intensity and transmission fixed in this experiment.

\begin{figure*}[ht]
\centering
\includegraphics[width=1\linewidth]{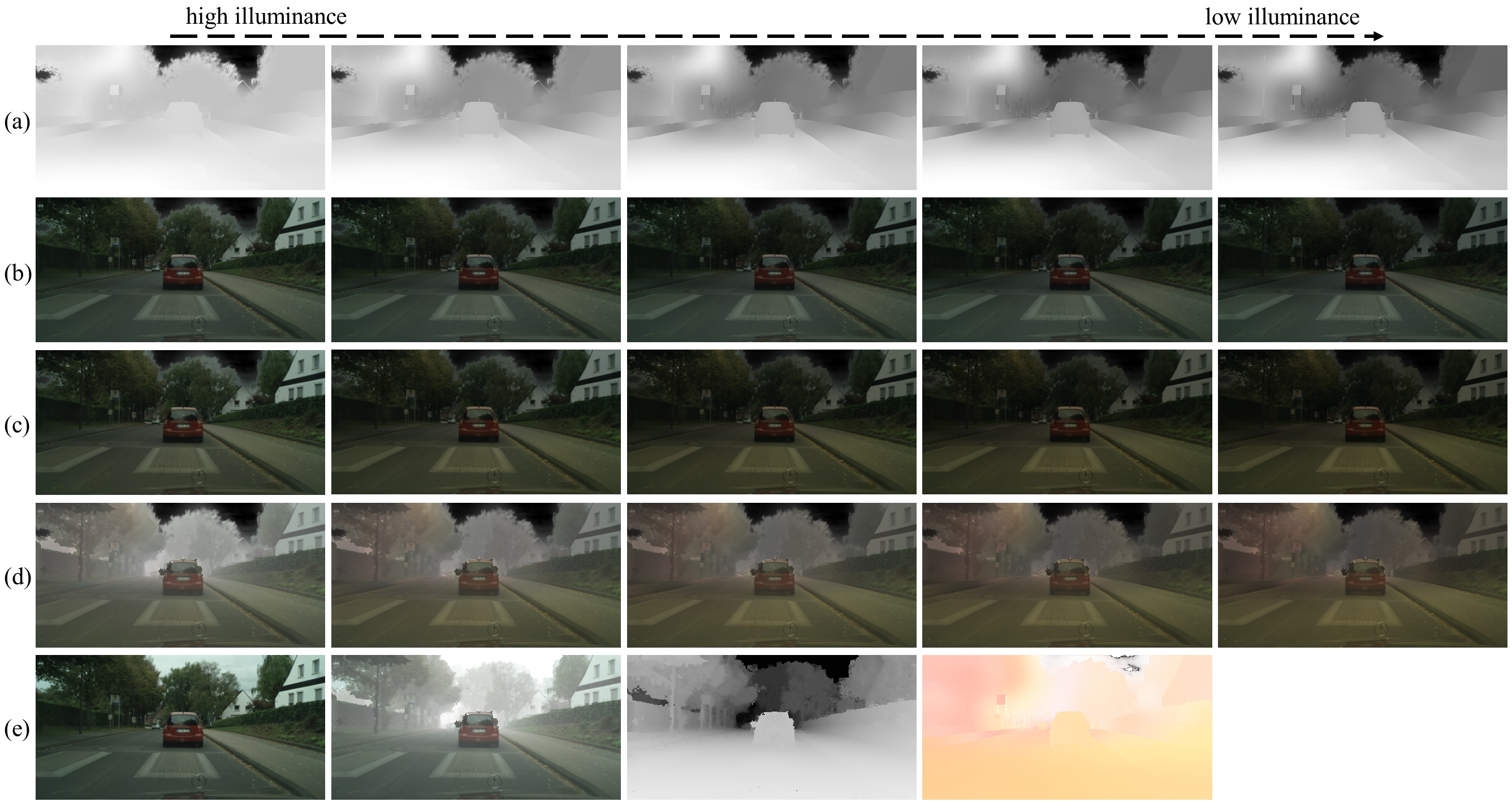}
\caption{(a) The illuminance maps at different illumination levels, $i.e.$, $\beta_l = 0.02,0.04,...,1$ (see Section~3.2 in the paper.). (b) The synthetic nighttime images without color cast generated by 3R. (c) The synthetic nighttime images with color cast generated by 3R. (d) The synthetic nighttime hazy images generated by 3R. (e) From left to right, the original daytime clear image, the daytime hazy image, the transmission map, and the color cast map.}
\label{fig:syn_l}
\end{figure*}

\textbf{Illuminance Intensity}
As shown in Figure~\ref{fig:syn_l}, with the increasing of $\beta_l$ from 0.02 to 1, the illuminance intensity becomes lower (Figure~\ref{fig:syn_l}(a)), and the nighttime (hazy) images (Figure~\ref{fig:syn_t}(b)$\sim$(d)) become darker accordingly. We kept the light colors and haze transmission (Figure~\ref{fig:syn_l}(e)) fixed in this experiment.

\subsubsection{Visual results on Virtual KITTI generated by 3R}
\label{subsec:virtualKitti}
\begin{figure*}[ht]
\centering
\includegraphics[width=1\linewidth]{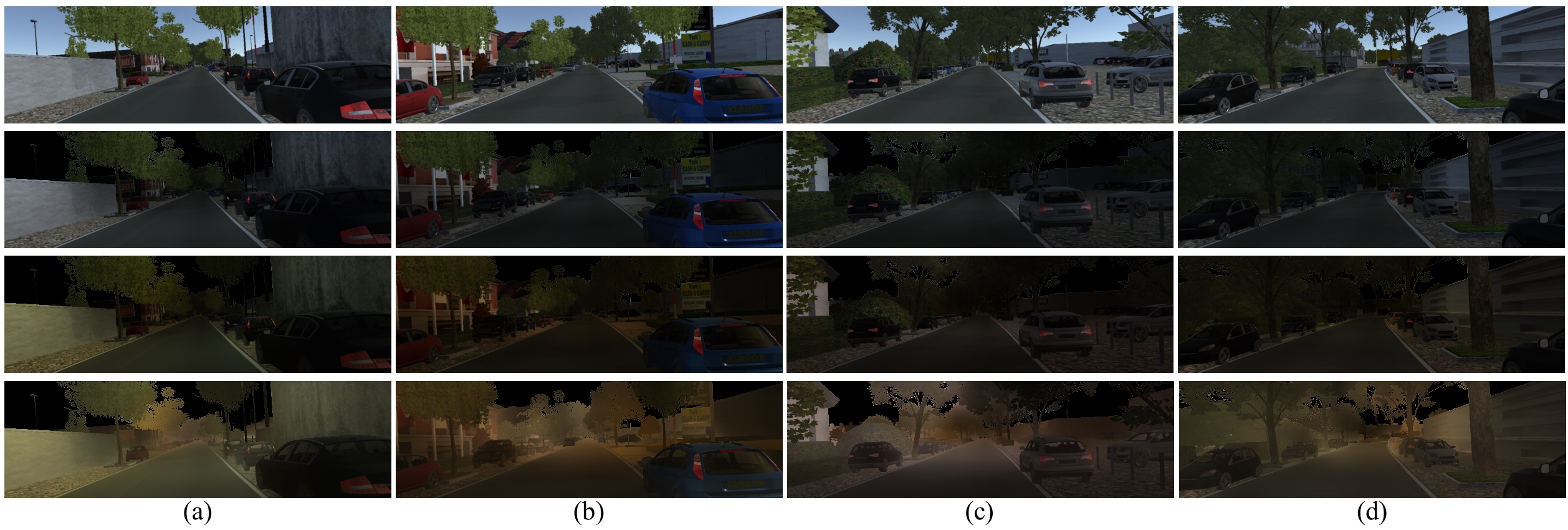}
\caption{(a) Daytime clear images from the Virtual KITTI dataset \cite{gaidon2016virtual}. (b) The synthetic nighttime images without color cast generated by 3R. (c) The synthetic nighttime images with color cast generated by 3R. (d) The synthetic nighttime hazy images generated by 3R.}
\label{fig:syn_virtualkitti}
\end{figure*}

Our 3R method can also generate nighttime hazy images based on the Virtual KITTI dataset \cite{gaidon2016virtual}. 
It is noteworthy that nighttime hazy imaging condition is not available in the dataset. Some results are shown in Figure~\ref{fig:syn_virtualkitti}, which look reasonable. Considering that the original images in the Virtual KITTI dataset are rendered by a 3D game engine where the textures are not so realistic, we do not include them in our benchmark. In the future, we can try to synthesize images on the original KITTI images \cite{geiger2012we} by estimating the absent depth maps or semantic labels.

\begin{figure*}[ht]
\centering
\includegraphics[width=1\linewidth]{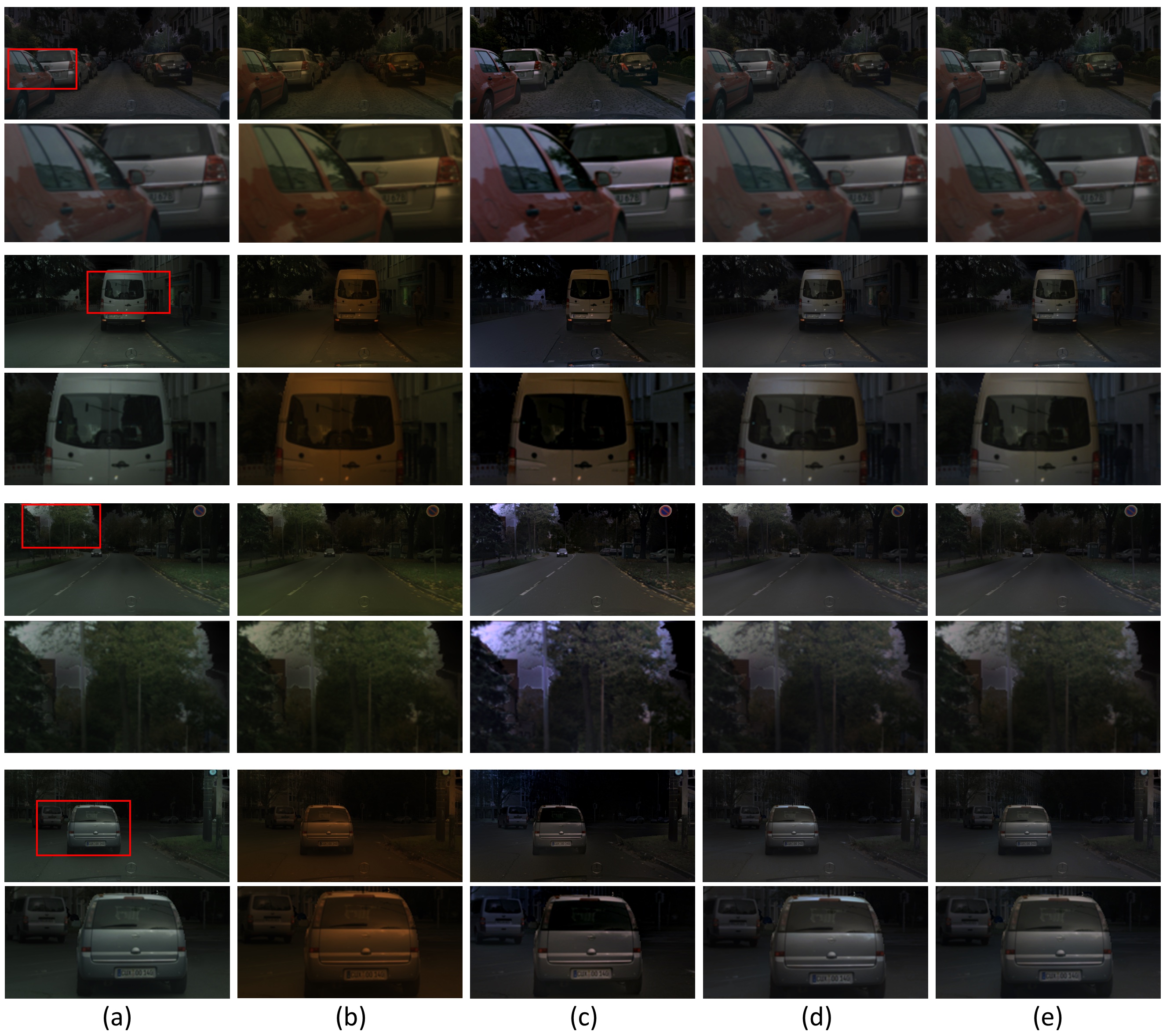}
\caption{(a)-(b) The synthetic nighttime images without/with color cast generated by 3R. (c) GS \cite{li2015nighttime}. (d) MRP \cite{zhang2017fast}. (e) Our OSFD. The even rows show the close-up views of the red rectangles. }
\label{fig:colorremoval_cityscapes}
\end{figure*}

\subsection{Comparison with state-of-the-art dehazing methods}
\subsubsection{Visual comparison for color cast removal}
\begin{figure*}[ht]
\centering
\includegraphics[width=0.65\linewidth]{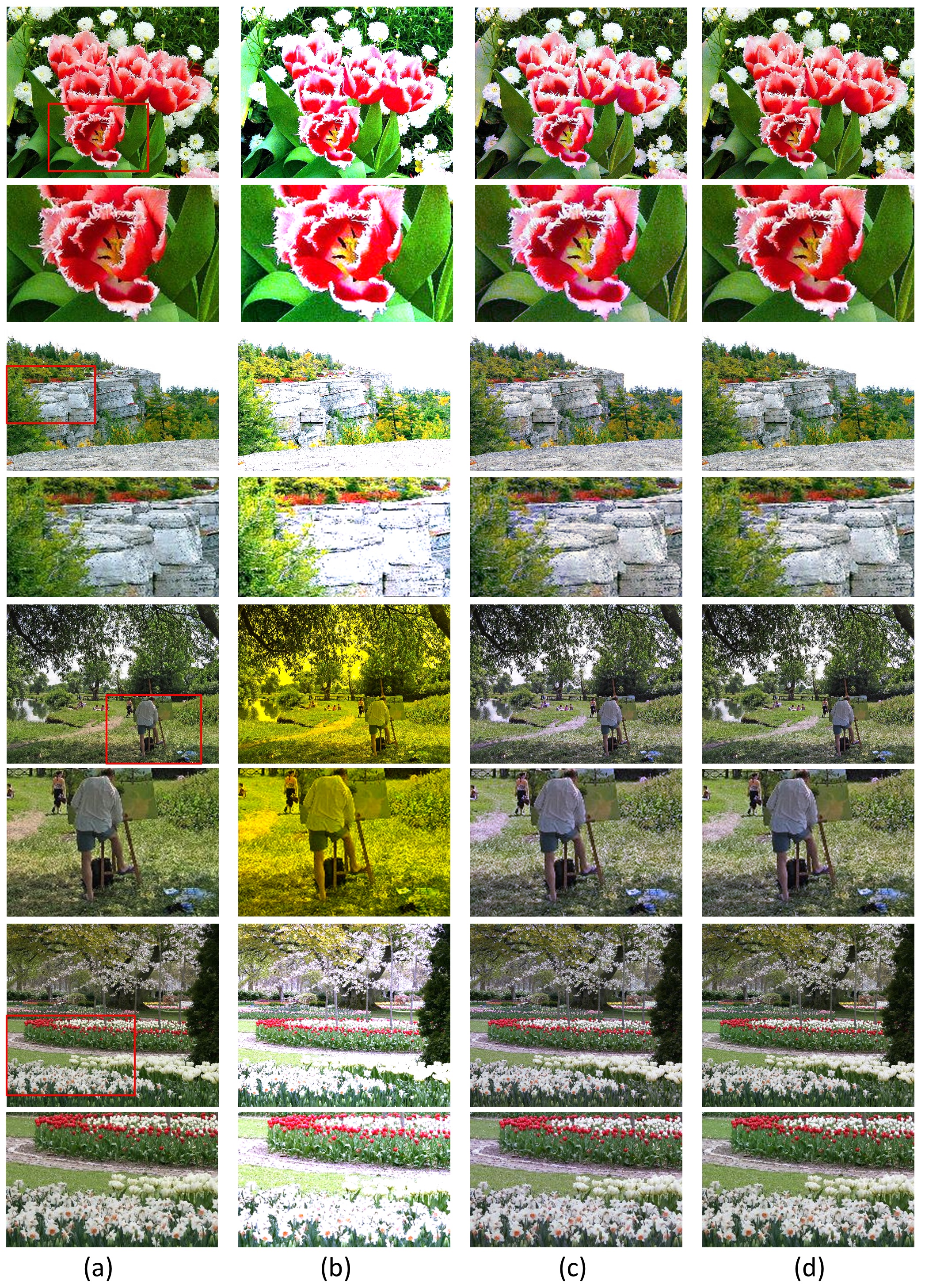}
\caption{(a) The daytime clear images from the DCRW dataset (see Section~3.3). (b) GS \cite{li2015nighttime}. (c) MRP \cite{zhang2017fast}. (d) Our OSFD. The even rows show the close-up views of the red rectangles. }
\label{fig:colorremoval_dcrw}
\end{figure*}

We also conducted an experiment to compare different methods for color cast removal as described in Section~5.2 and Table~4 in the paper. In this part, we present some visual results in Figure~\ref{fig:colorremoval_cityscapes} and \ref{fig:colorremoval_dcrw}. Since OSFD adaptively chooses the best scale to estimate the color cast, it can handle the monochromatic areas such as cars and trees in Figure~\ref{fig:colorremoval_cityscapes}, and avoid the whitish effect or residual color cast in MRP's results \cite{zhang2017fast}. The results on the daytime clear images without color cast also demonstrate the superiority of OSFD for retaining intrinsic image colors.

\subsection{Hyper-parameter settings}
\subsubsection{Hyper-parameter settings for ND-Net}
In this part, we present the brief ablation studies on the proposed ND-Net including training epochs, input image size, batch size, and losses. The results are summarized in Table~\ref{tab:ablationReside}. As can be seen, using a large input image size, training for a long time, as well as employing the perceptual loss can improve the dehazing results. We choose the setting listed in the last row as the default one in the paper. 

\begin{table}[htbp]
  \centering
  \caption{Hyper-parameter setting studies on the NHR dataset. E: training epochs; B: batch Size; S: input image size; MSE: MSE loss; MSE+PL: MSE and Perceptual loss}
    \begin{tabular}{lcccc}
    \toprule
      Method    & PSNR\,($\uparrow$)  & SSIM\,($\uparrow$)  & CIEDE2000\,($\downarrow$) \\
    \midrule
    E=70, B=8, S=128, MSE & 26.54 & 0.9286 & 5.82 \\
    E=70, B=8, S=256, MSE & 27.18 & 0.9312 & 5.05 \\
    E=300, B=8, S=256, MSE & \textbf{28.95} & 0.9413 & 4.30 \\
    E=300, B=8, S=256, MSE+PL & 28.74 & \textbf{0.9465} & \textbf{4.02} \\
    \bottomrule
    \end{tabular}%
  \label{tab:ablationReside}%
\end{table}%

\subsubsection{The number of scales in OSFD}
\begin{figure*}[ht]
\centering
\includegraphics[width=0.85\linewidth]{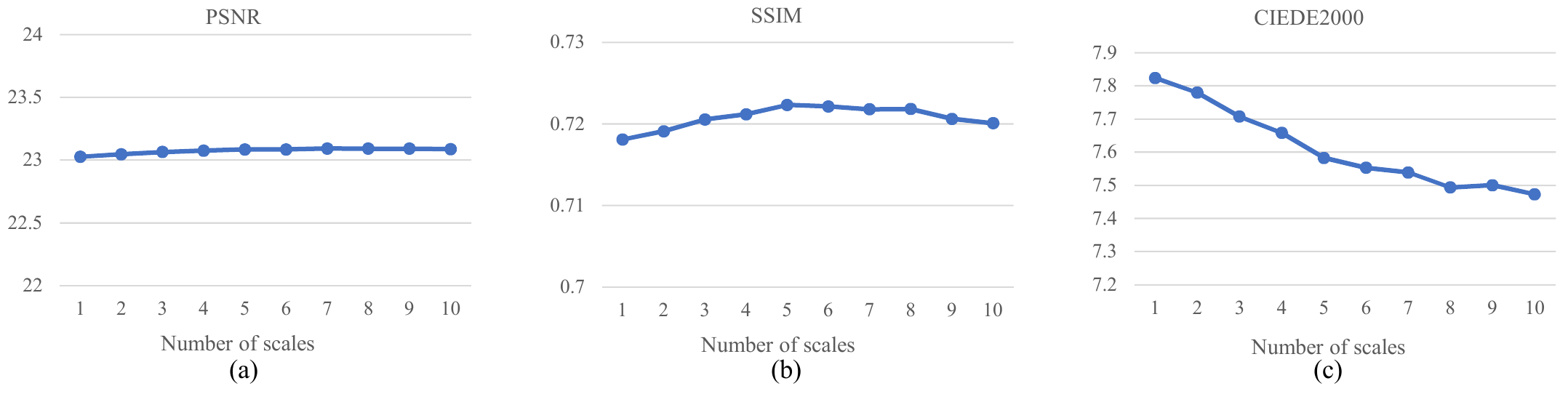}
\caption{Results of OSFD using different number of scales on the NHC-L dataset. (a) PSRN. (b) SSIM. (c) CIEDE2000. }
\label{fig:paramter}
\end{figure*}

We conducted a parameter study on the number of scales used in OSFD. The results are reported on the NHC-L dataset as shown in Figure~\ref{fig:paramter}. Using more scales, OSFD achieves lower CIEDE2000 scores, implying better color cast removal results. There is a steep drop at the beginning (1$\sim$5 scales), and then the performance saturates. Considering that OSFD is computationally efficient and the color artifacts in the dehazed results are annoying, we choose 10 scales for OSFD in other experiments to obtain better visual results.

\begin{table*}[htbp]
  \centering
  \caption{Standard deviations of the dehazing results on the NHC dataset.}
    \begin{tabular}{lccc|ccc|ccc}
    \toprule
    & \multicolumn{3}{c}{NHC-L} & \multicolumn{3}{c}{NHC-M} & \multicolumn{3}{c}{NHC-D}\\
    \hline
      Method    & PSNR\,($\uparrow$)  & SSIM\,($\uparrow$)  & CIEDE2000\,($\downarrow$) & PSNR\,($\uparrow$)  & SSIM\,($\uparrow$)  & CIEDE2000\,($\downarrow$) & PSNR\,($\uparrow$)  & SSIM\,($\uparrow$)  & CIEDE2000\,($\downarrow$)\\
    \midrule
    NDIM \cite{zhang2014nighttime}  & 0.0113 & 0.0010 & 0.0363 & 0.0136 & 0.0019 & 0.0524 & 0.0059& 0.0013 & 0.0123\\
    GS \cite{li2015nighttime} & 0.0172 & 0.0015 & 0.0268 & 0.0228 & 0.0014 & 0.0247 & 0.0146& 0.0011 & 0.0309\\
    MRPF \cite{zhang2017fast} & 0.0144 & 0.0007 & 0.0131 & 0.0131 & 0.0006 & 0.0139 & 0.0051& 0.0007& 0.0151\\
    MRP \cite{zhang2017fast} & 0.0031 & 0.0004 & 0.0091 & 0.0071 & 0.0005 & 0.0194 & 0.0024& 0.0009& 0.0174\\
    \textbf{OSFD} & 0.0072 & 0.0006 & 0.0134 & 0.0085 & 0.0008 & 0.0178 & 0.0016& 0.0010& 0.0196\\
    \bottomrule
    \end{tabular}%
  \label{tab:std}%
\end{table*}%

Since we augmented the Cityscapes by 5 times when generating the NHC dataset (see Section~3.3), the results listed in Table~2 in the paper are obtained by averaging the scores from all five tests. The corresponding standard deviations are listed in Table~\ref{tab:std}. As can be seen, the standard deviations are very low. Since we carried out each evaluation on all the 550 images, the score from each evaluation is stable even if we changed the light colors randomly during each augmentation.

\bibliographystyle{ACM-Reference-Format}
\balance
\bibliography{2020MM_ND}


\end{document}